\documentclass[10pt,twocolumn,letterpaper]{article}
\usepackage{cvpr}     

\usepackage[accsupp]{axessibility}
\usepackage{float}
\usepackage{url}
\usepackage{xcolor}
\usepackage{color, colortbl}
\usepackage{graphicx}
\usepackage{multirow}
\usepackage{threeparttable}
\usepackage{bbding}
\usepackage{tabularx}
\usepackage{makecell}
\usepackage{subcaption}
\usepackage[skins]{tcolorbox}
\usepackage{wrapfig}
\usepackage{longtable}
\usepackage{amssymb} 

\usepackage{tabularx}
\usepackage{algorithm}
\usepackage{algorithmic}
\usepackage{listings}
\usepackage{pifont}
\tcbuselibrary{breakable} 
\usepackage{enumitem}

\usepackage[accsupp]{axessibility}
\definecolor{cvprblue}{rgb}{0.21,0.49,0.74}
\usepackage[pagebackref,breaklinks,colorlinks,allcolors=cvprblue]{hyperref}
\definecolor{mydarkgreen}{rgb}{0.2,0.7,0.2}

\title{VideoChat-M1: Collaborative Policy Planning for Video Understanding \\ via Multi-Agent Reinforcement Learning}

\author{Boyu Chen\textsuperscript{\rm1,\rm2,\rm3 \thanks{Equal contribution.}}, Zikang Wang\textsuperscript{\rm4,\rm6 \footnotemark[1]}, Zhengrong Yue\textsuperscript{\rm6 \footnotemark[1]}, Kainan Yan\textsuperscript{\rm1,\rm2 \footnotemark[1]}, Chenyun Yu\textsuperscript{\rm5 \footnotemark[2]}, \\ Yi Huang\textsuperscript{\rm3 \footnotemark[2]}, Zijun Liu\textsuperscript{\rm8}, Yafei Wen\textsuperscript{\rm3}, Xiaoxin Chen\textsuperscript{\rm3}, Yang Liu\textsuperscript{\rm4,\rm7,\rm8}, Peng Li\textsuperscript{\rm7,\rm8 \footnotemark[2]}, Yali Wang\textsuperscript{\rm1,\rm4 \thanks{Equal corresponding author.}}\\
\textsuperscript{\rm1}Shenzhen Key Lab of Computer Vision and Pattern Recognition, Shenzhen Institutes of \\ Advanced Technology, Chinese Academy of Sciences\\
\textsuperscript{\rm2}School of Artificial Intelligence, University of Chinese Academy of Sciences \\
\textsuperscript{\rm3}VIVO AI Lab,
\textsuperscript{\rm4}Shanghai Artificial Intelligence Laboratory,\\
\textsuperscript{\rm5}Shenzhen Campus of Sun Yat-sen University,
\textsuperscript{\rm6}Shanghai Jiao Tong University, \\
\textsuperscript{\rm7}Institute for AI Industry Research (AIR), Tsinghua University,\\
\textsuperscript{\rm8}Dept. of Comp. Sci. \& Tech., Institute for AI, Tsinghua University\\
}

\begin{document}
\maketitle
\begin{abstract}

Most of the multi-agent video understanding frameworks adopt static and non-learnable tool invocation mechanisms, which limit the discovery of diverse clues essential for robust perception and reasoning regarding temporally or spatially complex videos.
To address this challenge, we propose a novel \textbf{M}ulti-agent system for video understanding, namely \textbf{VideoChat-M1}.
Instead of using a single or fixed policy, we adopt a distinct Collaborative Policy Planning (CPP) paradigm with multiple policy agents, which comprises three key processes.
(1) Policy Generation: Each agent generates its unique tool invocation policy tailored to the user's query; (2) Policy Execution: Each agent sequentially invokes relevant tools to execute its policy and explore the video content;
(3) Policy Communication: During the intermediate stages of policy execution, agents interact with one another to update their respective policies.
Through this collaborative framework, all agents work in tandem, dynamically refining their preferred policies based on contextual insights from peers.
Moreover, we equip our CPP paradigm with Multi-Agent Reinforcement Learning (MARL).
Consequently, policy agents can be jointly optimized to enhance the performance, guided by both the final answer reward and intermediate collaborative process feedback.
Extensive experiments demonstrate that VideoChat-M1 achieves SOTA performance across eight benchmarks on four tasks. 
Notably, on LongVideoBench, our method outperforms Gemini 2.5 pro by 3.6\% and GPT-4o by 15.6\%.

\end{abstract}

\begin{figure}[h]
    \centering
    \vspace{-1em}

    \includegraphics[width=0.45\textwidth]{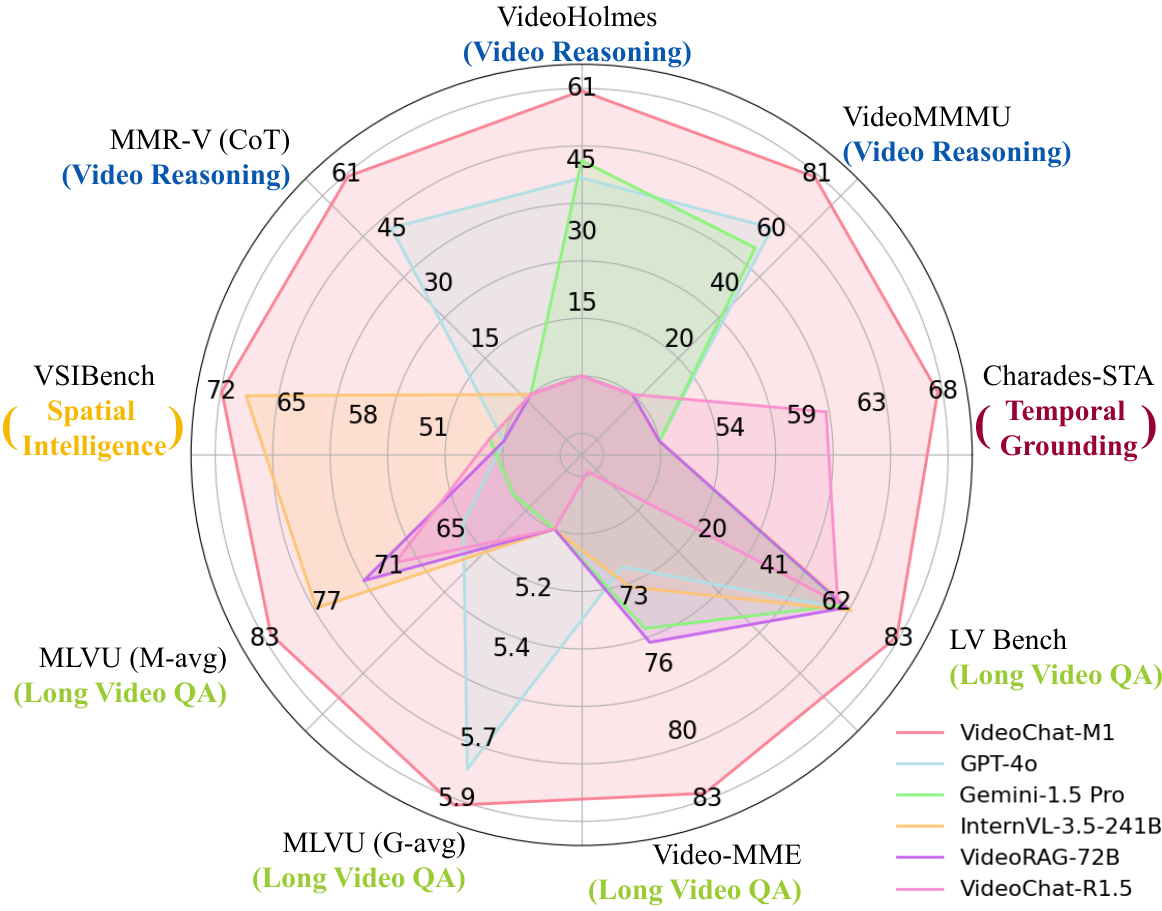}
    \vspace{-1em}
    \caption{
        \textbf{Comparison with SOTA.} VideoChat-M1 outperforms closed-source models (including GPT-4o) and open-source models (including InternVL-3.5-241B) in mainstream video tasks.
    }
    \label{fig:sota}
    \vspace{-2em}
\end{figure}

\begin{figure*}
    \centering
    \vspace{-1.2em}
    
    \includegraphics[width=\linewidth]{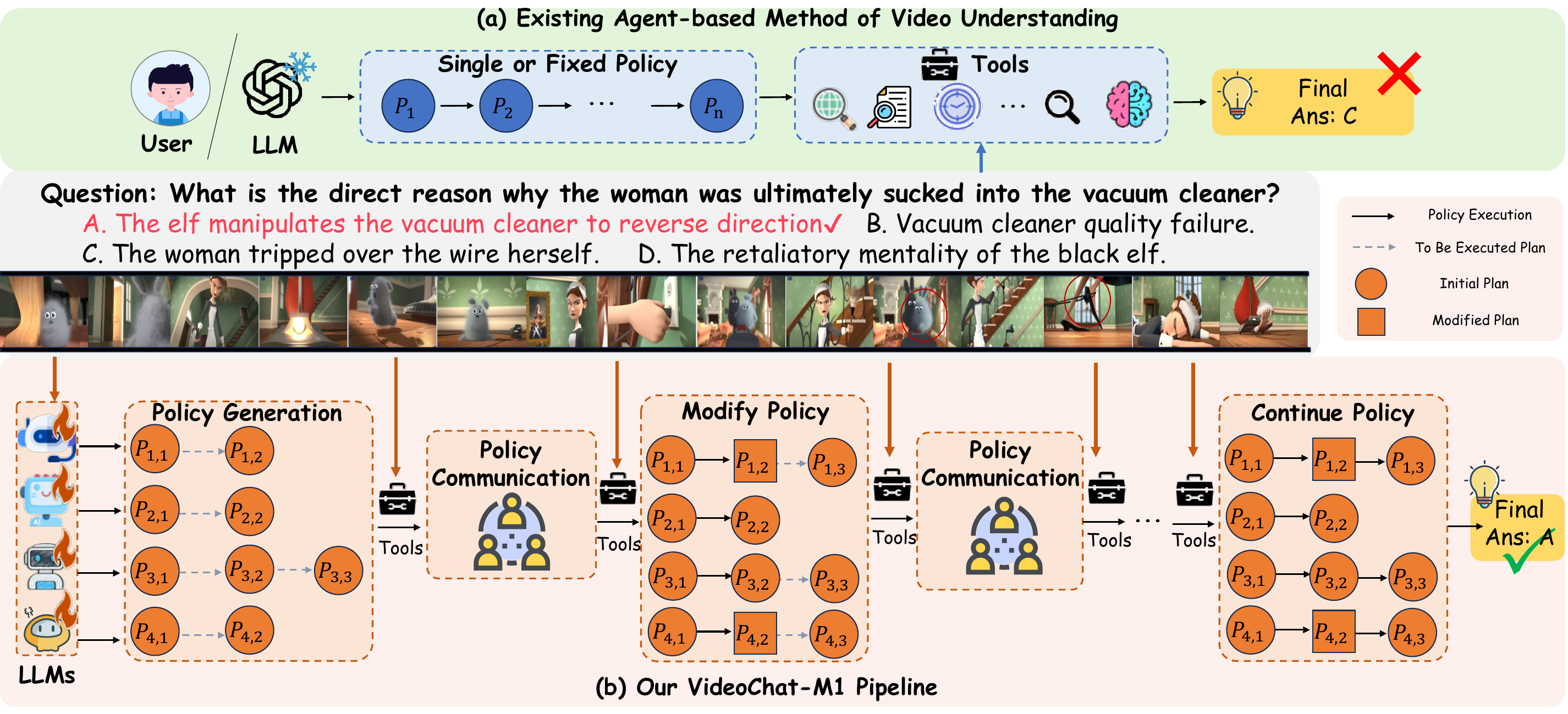}
    \vspace{-1.2em}
    
    \caption{Architecture and Working Mode Comparison (Existing Agent-based Method vs. Our VideoChat-M1). While prior methods rely on a fixed policy, VideoChat-M1 introduces a collaborative multi-agent policy planning pipeline that generates, executes, communicates and refines plans iteratively, enabling more adaptive and accurate long-video reasoning.}
    
    \label{fig:motivation}
    \vspace{-1.8em}
\end{figure*}

\section{Introduction}
\label{sec:intro}
Video understanding is a critical topic in computer vision~\cite{videoinedu,videoinmed,videoinent}.
Recent advancements in this field have been predominantly driven by Multimodal Large Language Models (MLLMs)~\cite{seed1.5vl,deepseek-r1,gpt4o,gemini2.5,Qwen2.5-VL,wang2025internvl35,eagle_2.5,longvila}. However, most MLLMs excel at processing short video clips while struggling to understand videos with long temporal contexts and/or complex spatial structures~\cite{gpt4o,gemini2.5,seed1.5vl,deepseek-r1,Qwen2.5-VL,wang2025internvl35}.
Recently, agent-based frameworks have shown significant potential to overcome these limitations in video understanding~\cite{videoagentmemory,videorag,videotree,vca,chen2025lvagent,wang2025videochat,videoagentrepeat}. 
Instead of directly feeding massive video frames into MLLMs, these frameworks enhance video understanding by invoking various tools to extract key video clues, either in an off-the-shelf~\cite{videotree,videoagentmemory} or iterative~\cite{wang2025videochat,videoagentrepeat,VITAL,videoexplorer,reagent-V,vca}
manner. 
Nevertheless, the tool invocation policy in existing agent-based frameworks is straightforward and fixed as shown in Fig~\ref{fig:motivation} (a), i.e., they adhere to pre-defined rules for tool selection and invocation during video understanding, without adaptive learning.
Such ad-hoc policies inherently prevent them from identifying, tracking, and summarizing rich clues on diverse temporal scales, leading to suboptimal perception and reasoning capabilities for complex videos~\cite{seed1.5vl,deepseek-r1,gpt4o,gemini2.5,Qwen2.5-VL,wang2025internvl35}.

To address this core challenge,
we introduce \textbf{VideoChat-M1},
a novel \textbf{M}ulti-agent system for multiple video understanding tasks.
Unlike frameworks with predefined tool policies, VideoChat-M1 introduces a Collaborative Policy Planning (CPP) paradigm where agents autonomously generate and adaptively update their policies for better video understanding. As shown in Fig~\ref{fig:motivation}(b), CPP involves three core stages: policy generation (where agents formulate tool invocation strategies), policy execution, and policy communication.
Subsequently, during policy execution process, each agent implements its policy progressively by using relevant tools to discover video clues.
To boost policy effectiveness and robustness, we further integrate policy communication into the execution process: after each step of execution, each agent receives video clues from peers.
Through this concise communication, agents synthesize contextual information from others and decide whether to refine their original policies into more optimal ones.
Via the CPP paradigm, all agents in our VideoChat-M1 framework collaborate to generate diversified tool-invocation policies, enabling the extraction of richer video clues to deeply understand complex queries or questions in videos.

To ensure the robustness and effectiveness of VideoChat-M1,
we equip the CPP paradigm with a concise Multi-Agent Reinforcement Learning (MARL) method.
To our best knowledge, this is the first policy learning framework that supports joint RL training among multiple agents for video understanding. 
Specifically, we treat each agent as a policy model and design three types of rewards for their joint optimization.
First, we use each agent's query answer and format constraints as rewards to encourage correct responses and penalize incorrect ones.
More critically, we employ an LLM as the reward model to evaluate the intermediate collaboration process, i.e., rewarding agents that generate superior policies via the CPP paradigm and penalizing those with inferior ones.
Guided by these rewards, we adapt Group Relative Policy Optimization (GRPO) to optimize the entire VideoChat-M1 agent group, creating a more effective policy planning to improve video understanding.

Finally, we evaluated VideoChat-M1 on 8 challenging benchmarks spanning
long video QA,
video reasoning,
spatial intelligence,
and 
temporal grounding.
Extensive experiments demonstrate that our VideoChat-M1 method achieves exceptional performance across all tasks, outperforming both closed-source and open-source baselines.
Notably,
For the long video question answering task on LongVideoBench~\cite{wu2024longvideobench}, we outperform GPT-4o~\cite{gpt4o} by 15.6\%. 
For video reasoning on VideoMMMU~\cite{hu2025videommmu}, our 37B agent group delivers results comparable to Qwen3-VL-235B~\cite{Qwen2.5-VL} while using only 15\% of the model parameters.
In the spatial intelligence task on VSIBench~\cite{yang2025thinking}, our model exceeds Gemini 1.5 Pro~\cite{gemini15} by 26.5\%.
For the temporal grounding task on Charades-STA~\cite{gao2017tall}, we achieve a 3.0\% improvement over Seed 1.5VL~\cite{seed1.5vl}. 
Our key contributions are summarized as follows:
\begin{itemize}
    \item We propose VideoChat-M1, the first multi-agent framework for video understanding that replaces the conventional single, fixed policy with a Collaborative Policy Planning (CPP) paradigm, enabling agents to dynamically generate and adapt tool-use strategies through multi-agent policy communication.
    \item  We introduce a pioneering Multi-Agent Reinforcement Learning (MARL) method to optimize the collaborative process. It uniquely employs a hybrid reward system that evaluates both the final answer accuracy and the intermediate quality of multi-agent collaboration.
    \item Extensive experiments show that VideoChat-M1 achieves SOTA performance on eight challenging benchmarks. It significantly outperforms leading models such as GPT-4o and Gemini 1.5 Pro, while exhibiting superior parameter efficiency compared to substantially larger models.
\end{itemize}

\section{Related Work}
\label{sec:related}

\noindent\textbf{Video Understanding.}
Understanding videos is a major challenge for Vision-Language Models (VLMs). Early efforts improved single-model performance through architecture scaling~\cite{chen2022low, videochat, llavavideo, internvl2.5}, context window extension~\cite{lwm, gemini, longllava}, token compression~\cite{moviechat, videollama, llamavid}, or reinforcement learning~\cite{longrl, li2025videochat-r1}.
However, these approaches struggle with precise retrieval because a single agent cannot easily manage perception, retrieval, and synthesis simultaneously
To address this issue, subsequent work has equipped a single agent with retrieval~\cite{zhao2024longagent}, memory~\cite{videoagentmemory, mallm}, and search tools~\cite{searchlvlm, videotree, timesuite}, but their general-purpose design limits effective integration and reasoning.  
Unlike systems such as LVAgent~\cite{chen2025lvagent} that rely on static, untrained collaboration, we introduce a trainable multi-agent framework for dynamic adaptability across diverse video tasks.

\noindent\textbf{Multi-Agent Reinforcement Learning.} Recent advancements in LLMs have driven the development of multi-agent systems, which follow two paradigms. First, training-free systems (e.g., CAMEL~\cite{li2023camel} and MetaGPT~\cite{hong2023metagpt}) rely on engineered logic and fixed roles. However, their static text-centric design often fail to handle dynamic multi-modal tasks like long-form video understanding.
Second, RL-based paradigm train collaborative policies by optimizing agent behaviors~\cite{zhuge2024gptswarm,  liao2025marft, park2025maporl, CURE,vragent,pca,chen2025super,yue2025uniflow,wang2025videochat,chen2025g, chen2025top} or interaction architectures~\cite{maas, gdesigner, motwani2024malt}. Despite growing sophistication~\cite{MLPO, wan2025rema, gao2025flowreasoner, wei2025lero, MATPO}, these methods remain confined to unimodal text domains, overlooking temporal and perceptual challenges specific to video.
Existing RL methods struggle to co-train heterogeneous agents, limiting their synergy. To address this, we draw inspiration from successful multi-agent GUI pipelines~\cite{he2025enhancing} to introduce VideoChat-M1, a novel framework that enables joint training of diverse agents for complex multi-modal tasks.

\section{Methodology}
\label{sec:method}

This section details the VideoChat-M1 framework, introducing its Collaborative Policy Planning (CPP) paradigm, followed by the Multi-Agent Reinforcement Learning (MARL) method that enhances its effectiveness.

\subsection{Collaborative Policy Planning Pipeline (CPP)}

As noted earlier,
existing agent-based frameworks primarily adopt a single and fixed tool invocation policy,
resulting in suboptimal performance on long-form and complex videos.
Therefore, we propose a distinct CPP paradigm for enhancing video understanding, where multiple agents collaborate to dynamically refine tool-invocation policies. For clarity, let $\mathcal{Q}$ denote a user query for a video $\mathcal{V}$ (a question about specific video content). Our CPP paradigm comprises a set of policy agents $\mathcal{G} = \{\mathcal{G}_i\}$,
a set of video perception tools $\mathcal{T} = \{\mathcal{T}_j\}$,
and a shared memory buffer $\mathcal{M} = \{\mathcal{M}_i\}$.
This buffer records key historical information from all agents
(e.g., initial policy plan, intermediate answer) and supports policy updates through agent communication. Further details are provided in Appendix A.2.
Fig~\ref{fig:cpp} illustrates the CPP workflow, an iterative collaborative framework where each policy agent autonomously executes three core phases: policy generation, policy execution, and policy communication. During each execution step, all agents share the group’s state, key video clues, and decision-making information via the shared memory buffer, dynamically optimizing tool-invocation decisions for the next step. Through the CPP pipeline, agents collaborate to generate diversified tool-invocation plans, extracting richer video clues to deeply understand complex video queries.

\begin{figure*}[t]
    \centering
    \includegraphics[width=\linewidth]{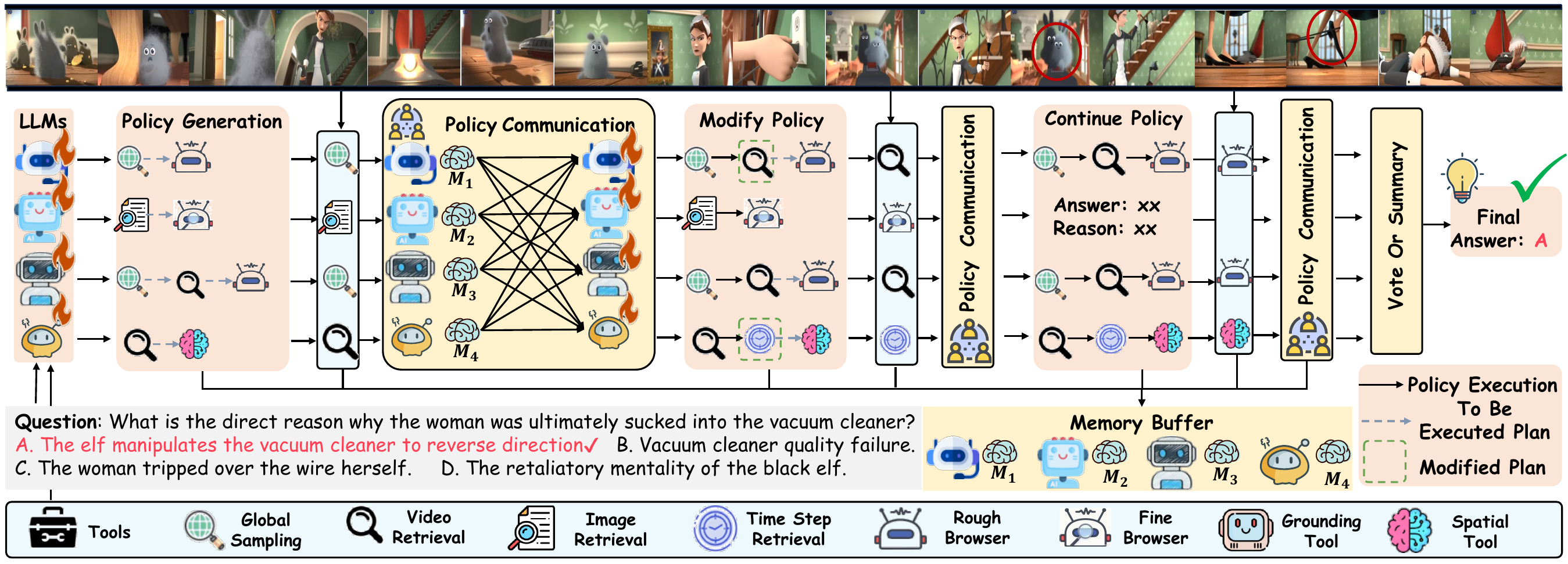}
    \vspace{-2em}
    
    \caption{The workflow of Collaborative Policy Planning (CPP) in the Reasoning Phase. 
    Multiple agents independently generate initial plans, communicate to exchange reasoning states, and iteratively refine their policies using different tools. Through repeated rounds of communication and plan updates, the agents collectively vote or summarize to produce a reliable final answer.
    }
    
    \label{fig:cpp}
    \vspace{-1.5em}
\end{figure*}
\subsubsection{Policy Generation}

Our CPP pipeline begins with a policy generation stage, 
where the core video understanding task is sequentially decomposed into smaller and manageable sub-tasks. 
Specifically, each agent $\mathcal{G}_i$ generates an initial policy, which specifies an explicit solution to address the user query $\mathcal{Q}$ by invoking a sequence of tools from toolkit $\mathcal{T}$.
This process is formally defined as:
\begin{equation}
\mathcal{P}_i = \mathcal{G}_i(\mathcal{Q}, ~\mathcal{T}),
\label{eq:gen}
\setlength\abovedisplayskip{1pt}
\setlength\belowdisplayskip{1pt}
\end{equation}
where $\mathcal{P}_i=\{\mathcal{P}_{i,1}\rightarrow\mathcal{P}_{i,2}\rightarrow...\rightarrow\mathcal{P}_{i,N}\}$ denotes the initial policy of agent $\mathcal{G}_i$,
and
$\mathcal{P}_{i,N}$ is the $N$-th policy step that specifies a certain tool in $\mathcal{T}$ for analyzing the input video.

\subsubsection{Policy Execution}

After generating the initial policy $\mathcal{P}_i$, agent $\mathcal{G}_i$ proceeds with execution.
Specifically,
at the $n$-th policy step,
the agent utilizes $\mathcal{P}_{i,n}$ to retrieve the corresponding tool from the toolkit $\mathcal{T}$.
It then employs this tool to analyze the input video $\mathcal{V}$ to obtain the intermediate answer at the step $n$,
based on the $(n-1)$-th step answer:
\begin{equation}
\setlength\abovedisplayskip{1pt}
\setlength\belowdisplayskip{1pt}
\mathcal{A}_{i,n} = \mathcal{P}_{i,n}(\mathcal{V},~\mathcal{T},~\mathcal{A}_{i,n-1}),
\label{eq:ans}
\end{equation}
where
$\mathcal{A}_{i,n}$ and $\mathcal{A}_{i,n-1}$ denote the $n$-th and $(n-1)$-th step answers for agent $\mathcal{G}_i$, respectively.
This process iterates until obtaining the final answer of user's query,
i.e.,
$\mathcal{A}_i=\{\mathcal{A}_{i,1}\rightarrow\mathcal{A}_{i,2}\rightarrow...\rightarrow\mathcal{A}_{i,N}\}$.

However, the initial policies of agents may lack reliability, executing such suboptimal policies may fail to generate satisfactory results.
Hence, we propose a policy communication stage integrated with policy execution, enabling dynamic updates to tool-invocation policies as needed.

\subsubsection{Policy Communication}

To enhance the robustness of tool-invocation policies for answering the user query,
we introduce policy communication with agent group collaboration.
After executing each step of their initial policies,
the agent team $\mathcal{G}$ generates intermediate results $\mathcal{A}$,
and stores them in a shared memory $\mathcal{M}$.
Subsequently,
each agent references its initial policy and the team's intermediate memory 
to determine whether to update its policies for subsequent steps, formulated as:
\begin{equation}
\setlength\abovedisplayskip{1pt}
\setlength\belowdisplayskip{1pt}
\mathcal{P}'_{i} = \mathcal{G}_i(\mathcal{Q}, \mathcal{T}, \mathcal{M}, \mathcal{P}_i), 
\label{eq:policy_comm}
\end{equation}
where
$\mathcal{P}'_{i}$ denotes the updated policy of agent $\mathcal{G}_i$.
If the current policy $\mathcal{P}_{i}$ remains optimal, 
 $\mathcal{P}'_{i}$ stays unchanged,
and the agent continues tool invocation based on the next step of $\mathcal{P}_{i}$.
Otherwise,
the agent revises $\mathcal{P}_{i}$ as $\mathcal{P}'_{i}$,
and executes the next steps of the updated policies.

Moreover, policy communication and execution are performed iteratively. This allows each agent to effectively leverage the team's intermediate results as historical experience and refine its own policy through multi-round communication during policy execution.
Once all tools are executed, each agent summarizes its prior results to generate an answer.
The final answer is determined by the query type: multiple-choice questions are resolved by majority voting, while open-ended and temporal grounding queries are consolidated by a designated agent, which is the best-performing model in the group (Qwen3-8B~\cite{yang2025qwen3}).

\subsection{Multi-Agent Reinforcement Learning (MARL)}

Unlike previous multi-agent video understanding frameworks (no training),
we propose MARL to guide the team of policy agents,
enhancing VideoChat-M1's adaptability and collaborative capabilities.
To the best of our knowledge, this is the first multi-agent policy learning framework designed to tackle complex video understanding tasks.
As a warm-up, the supervised fine-tuning (SFT) stage equips each agent with basic abilities to produce a high-quality initial policy plan.
Subsequently, the MARL pipeline trains the agent team to achieve effective collaboration.

\begin{figure*}[t]
    \centering
    \includegraphics[width=\linewidth]{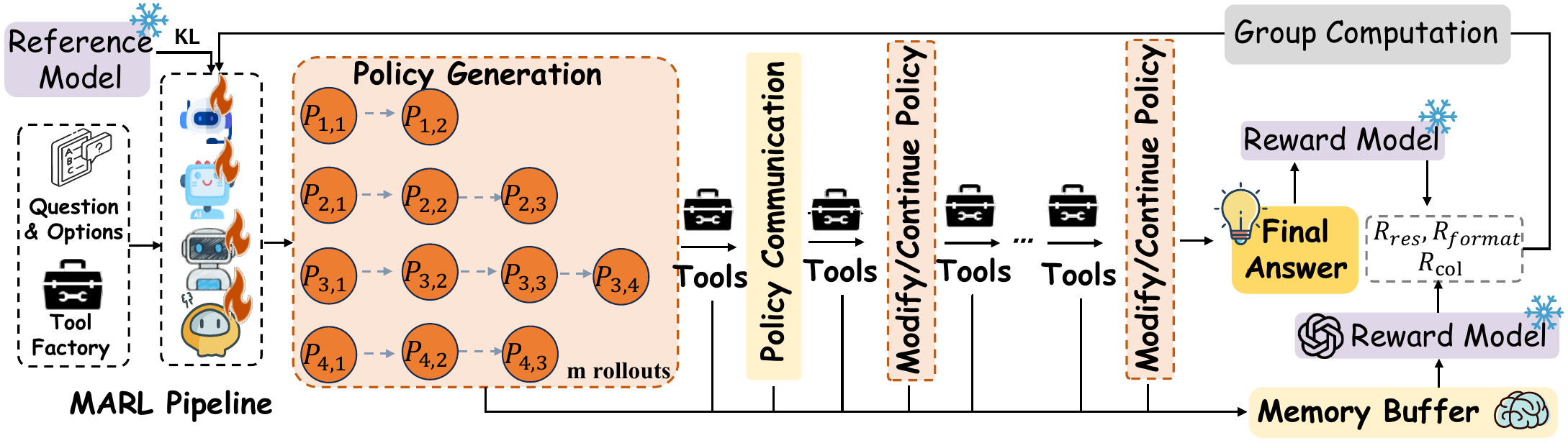}
    \vspace{-2.2em}
    
    \caption{Training the Agent Group using Our Multi-Agent Reinforcement Learning (MARL) Method. Agents generate policies, communicate, and iteratively refine them with tool feedback, while reward and reference models guide stable joint optimization.
    }
    \label{fig:marl}
    \vspace{-1.9em}
\end{figure*}

\subsubsection{Policy SFT}

We first construct a high-quality policy set from open-source video datasets (see Appendix A.1) for policy SFT. 
The input is the user query for the training video, and the output is the policy plan for answering the query.
Notably, since these open-source datasets lack pre-existing policy plans, we leverage our CPP paradigm with a high-performance team (i.e., GPT-4o and DeepSeek-R1) to automatically annotate policy plans for each training video. To guarantee annotation quality, we curate the policy data based on two core criteria: first, selecting annotated policy plans that yield the correct answer to the user's query for the training video; second, choosing plans that can be executed successfully to obtain the final answer without any policy modification. 
These criteria ensure the supervision signal comprises effective and efficient policy plans.
Using this policy plan dataset,
we fine-tune each agent in the team using 
the cross-entropy loss by maximizing the likelihood of generating the ground-truth plan.
This SFT phase enables each agent to master the basic capacity to generate a preferred policy plan in a structured output format, laying the foundation for collaborative learning in MARL.

\subsubsection{MARL}

To foster effective team collaboration, 
we introduce the MARL framework as shown in Fig~\ref{fig:marl} to further optimize the agent group.
Specifically,
we leverage three types of reward for this purpose,
i.e.,
$\mathcal{R} = \mathcal{R}_{res} + \mathcal{R}_{format} + \mathcal{R}_{col}$.

\begin{table*}[t]
\centering
\small
\setlength{\tabcolsep}{0.75mm}
\renewcommand{\arraystretch}{0.8}

\resizebox{\textwidth}{!}{

\begin{tabular}{l|c|ccc|cc|c|c|c|cccc|c}
\toprule

\multirow{3}{*}{\textbf{Model}} & \multicolumn{6}{c|}{\textbf{Long Video QA}} & \multicolumn{3}{c|}{\textbf{Video Reasoning}} & \multicolumn{4}{c|}{\textbf{Spatial Intelligence}} & \multicolumn{1}{c}{\begin{tabular}[c]{@{}c@{}} \textbf{Temporal}\\ \textbf{Grounding}  \end{tabular}} \\
\cmidrule(lr){2-7} \cmidrule(lr){8-10} \cmidrule(lr){11-14} \cmidrule(lr){15-15}

& \multirow{2}{*}{\begin{tabular}[c]{@{}c@{}} \textbf{LongVideo}\\ \textbf{Bench}  \end{tabular}} 
& \multicolumn{3}{c|}{\textbf{Video-MME}} 
& \multicolumn{2}{c|}{\textbf{MLVU}} 
& \multirow{2}{*}{\begin{tabular}[c]{@{}c@{}} \textbf{Video}\\ \textbf{Holmes}  \end{tabular}} 
& \multirow{2}{*}{\begin{tabular}[c]{@{}c@{}} \textbf{Video}\\ \textbf{MMMU}  \end{tabular}}
& \textbf{MMR-V} 
& \multicolumn{4}{c|}{\textbf{VSIBench}} 
& \textbf{Charades} \\
\cmidrule(lr){3-5} \cmidrule(lr){6-7} \cmidrule(lr){10-10}  \cmidrule(lr){11-15}

& 
& \textbf{M} & \textbf{L} & \textbf{Avg} 
& \textbf{M-avg} &  \textbf{G-avg} 
& 
& 
& \textbf{CoT} 
& \textbf{Dist} & \textbf{Dir} & \textbf{Order} & \textbf{Avg} 
& \textbf{m-IOU} \\
\midrule
\multicolumn{15}{c}{\textbf{\textit{Closed-source Large-sized MLLM}}} \\ \midrule
Gemini 2.5 Pro~\cite{gemini2.5}& \textcolor{blue}{78.7} & - &   -&   \textcolor{blue}{84.3} & - & - & - & 83.6 & - &   -&   -&   -& - & - \\

Gemini 1.5 Pro~\cite{gemini15} & 64.0 & \textcolor{blue}{74.3}& \textcolor{blue}{67.4}& 75.0 & - & - & \textcolor{blue}{45.7} & 53.9 & - & \textcolor{blue}{51.3}& \textcolor{blue}{46.3}& \textcolor{blue}{34.6}& \textcolor{blue}{45.4} & - \\ 

GPT-5-thinking& - & -& -& - & - & - & - & \textcolor{blue}{84.6} & - & - & - & - & - & -\\
GPT-4o~\cite{gpt4o} & 66.7 & 70.3 & 65.3& 71.9 & 64.6 & \textcolor{blue}{5.80} & 42.0 & 61.2 & \textcolor{blue}{46.1} & 37.0& 41.3& 28.5& 34.0 & - \\

OpenAI O3 & - & -& -& - & - & - & - & 83.3 & - & - & - & - & - & -\\

Seed 1.5VL~\cite{seed1.5vl} & 74.0 & - &   -&   77.9 & \textcolor{blue}{82.1} & - & - & 81.4 & - &   -&   -&   -& - & \textcolor{blue}{64.7} \\
\midrule

\multicolumn{15}{c}{\textbf{\textit{Open-source Large-sized MLLM}}} \\ \midrule
InternVL-3.5-241B~\cite{wang2025internvl35} & \textcolor{blue}{67.1} & - & - & 72.9 & 78.2& - & - & - & - & - & - & -& \textcolor{blue}{69.5} & - \\ 
Qwen3-VL-235B-Instruct~\cite{Qwen2.5-VL} & - & - & - & \textcolor{blue}{79.2} & \textcolor{blue}{84.3}& - & - & 74.7 & - & - & - & -& 62.6 & \textcolor{blue}{64.8} \\ 
Qwen3-VL-235B-Thinking~\cite{Qwen2.5-VL} & - & - & - & 79.0 & 83.8& - & - & \textcolor{blue}{80.0} & - & - & - & -& - & 63.5 \\
Qwen2.5-VL-72B~\cite{Qwen2.5-VL}                & 60.7 & - & - &  73.3  & 74.6& - & - & - &  40.4 & - & - & -& - & 50.9  \\
Qwen2-VL-72B~\cite{qwen2vl}                     & - & \textcolor{blue}{71.3} &   62.2&   71.2 &     - & - & - & - & \textcolor{blue}{40.4} &    -  &     - &      -& 36.1 & - \\
LLAVA-Video-72B~\cite{llavavideo}               &  64.9 & 68.9&   61.5&   70.6 & - & - & - & 49.7 & - & 42.4&   36.7&   48.6& 40.9 & -  \\
LLaVA-ov-72B~\cite{llavaonevision}              &  61.3 & 62.2 & 60.0 & 66.3 & 68.0 & - & - & 48.3 & - & 42.5 & \textcolor{blue}{39.9} & 44.6 & 40.2 & -  \\
VideoLLaMA2-72B~\cite{videollama2}              & - & 59.9&   57.6&   62.4 & 45.6& 3.78 & - & - & - & - & - & -& - & - \\
InternVL-2.5-78B~\cite{internvl2.5}             &  63.6 & 70.9&   \textcolor{blue}{62.6}&   72.1 & 75.7& - & - & - & - & - & - & -& - & - \\

InternVL-3-78B~\cite{zhu2025internvl3}          & 65.7 & - & - &  72.7  & 79.5& - & - & - & - & \textcolor{blue}{55.9}&   39.5&  \textcolor{blue}{54.5} & 48.4 & -  \\
InternVL-3.5-78B~\cite{wang2025internvl35}      & 65.7 & - & - &  70.9 & 77.0& - & - & - & - & - & - & -& 66.3 & - \\ 
Aria-28B~\cite{aria}                            & 64.2 & 67.0&   58.8&   67.6 & 72.3 & \textcolor{blue}{5.02} & - & 50.8 & - & - & - & -& - &  - \\
Oryx-34B~\cite{liu2024oryx}                     & - & 65.3&   59.3&   67.3 & 70.6 & - & - & - & - & - & - & -& - & -  \\

 \midrule
\multicolumn{15}{c}{\textbf{\textit{Open-source Medium-sized MLLM}}} \\ \midrule
VideoXL2-8B~\cite{videoxl-2}                    & 61.0 & - & - &  66.6      & 74.8 & - & - & - & - & - & - & -& - & 54.2 \\

Video-R1-7B~\cite{video-r1}                     & - & - & - & 61.4 & - & - & \textcolor{blue}{36.5} & - & - & - & - & -& 37.1 & -  \\
VideoChat-Flash-7B~\cite{videochatflash}        & 64.7 & - & \textcolor{blue}{55.4} & 65.3 & 74.7 & - & - & - & - & - & - & -& - & 48.0  \\

VideoChat2-7B~\cite{mvbench}                    & 39.3 & 37.0&   33.2&   39.5 & 47.9 &  3.81 & - & - & - & - & - & -& - &  - \\

VideoChat-R1-7B~\cite{li2025videochat-r1}       & - & - & - & - & - & - & 33.0 & - & - & - & - & -& - &  - \\
VideoChat-R1.5-7B~\cite{videochat-r1.5}         & 62.6 & - & - & 67.1 & 70.9 & - & - & 51.4 & - & - & - & -& - & 60.6 \\

InternVideo2.5-7B~\cite{wang2025internvideo2}             &  60.6 & -&   -&   65.1 & 72.8& - & - & - & - & - & - & -& - & - \\
InternVL-2.5-8B~\cite{internvl2.5}              & 60.0 & - & - & 64.2 & 68.9& - &  23.6 & - & - & - & - & -& -&  - \\
InternVL-3-8B~\cite{zhu2025internvl3}           & 58.8 & - & - &  66.3  & 71.4& - & 32.3 & - & \textcolor{blue}{32.9} & \textcolor{blue}{48.3}&   36.4&   \textcolor{blue}{35.4}& 42.1 &  - \\
InternVL-3.5-8B~\cite{wang2025internvl35}       & 62.1 & - & - &  66.0  & 70.2& - & - & - & - & - & - & -& 56.3 &  - \\
Qwen2-VL-7B~\cite{qwen2vl}                      & 55.6 & - & - & 63.3 & - & - &  27.8 & - & 32.4 & - & - & -& - & - \\

Qwen2.5-VL-7B~\cite{Qwen2.5-VL}                 & 56.0 & - & - &  65.1  & 70.2& - & - & - & - & - & - & -& 35.9 & 43.6 \\
Qwen3-VL-8B-Instruct~\cite{Qwen2.5-VL}                 & - & - & - & 71.4 & \textcolor{blue}{78.1}& - & - & \textcolor{blue}{65.3} & - & - & - & -& \textcolor{blue}{59.4} & 55.5 \\
LongVA-7B~\cite{longva}                         & 51.3 & 50.4&   46.2&   52.6 & - & \textcolor{blue}{4.33} & - & 24.0 & - & 33.1&   \textcolor{blue}{43.3}&   15.7& 29.2 & -  \\

LongVILA-7B~\cite{longvila}                     & 57.1 & 58.3&   53.0&   60.1 & - & - & - & - & - & - & - & -& - & -  \\
LongRL-7B~\cite{longrl}                         & 58.1 & \textcolor{blue}{63.2}&   55.2&   65.1 & - & - & - & - & - & - & - & -& - & - \\
LLaVA-Video-7B~\cite{llavavideo}                & 58.2 & - & - & 63.3 & 70.8 & 3.84 & - & 36.1 & 17.6 & 43.5&   42.4&   30.6& 35.6 &  - \\
Eagle-2.5-8B~\cite{eagle_2.5}                   & \textcolor{blue}{66.4} & - & - &  \textcolor{blue}{72.4}     & 77.6 & - & - & - & - & - & - & -& - & \textcolor{blue}{65.9} \\
ShareGPT4Video-8B~\cite{chen2024sharegpt4video} & 39.7 & 36.3&   35.0&   39.9 & 46.4 &  3.77 & - & - & - & - & - & -& - & - \\

 \midrule

\multicolumn{15}{c}{\textbf{\textit{Agent-based Methods}}} \\ \midrule
DeepVideoDiscovery~\cite{deepvideo}             & \textcolor{blue}{71.6} & - & 67.3 & - & - & - & - & - & - & - & - & -& - & - \\
DrVideo~\cite{drvideo}                          & - & - & - &   51.7 & - & - & - & - & - & - & - & -& - &  - \\
ReAgent-V-72B~\cite{reagent-V}                  & - & 72.3 & 72.9 & 75.1 & 74.2 & - & - & - & - & - & - & -& - & - \\ 
VCA~\cite{vca}                                  & 41.3 & - & - & - & - & - & - & - & - & - & - & -& - & - \\

VITAL-7B~\cite{VITAL}                           & - & - & 54.0 & 64.1 & - & - & - & - & - & - & - & -& - & \textcolor{blue}{59.9} \\
VideoChat-A1~\cite{wang2025videochat}            & 65.4 & 72.8&   65.0&   72.9 & \textcolor{blue}{76.2} & - & - & - & - & - & - & -& - & - \\
VideoExplorer-14B~\cite{videoexplorer}          & - & - & - & - & 55.4 & - & - & - & - & - & - & -& - & -  \\

VideoExplorer-39B~\cite{videoexplorer}          & - & - & - & - & 58.6 & - & - & - & - & - & - & -& - &  - \\

VideoRAG-72B~\cite{videorag}                    & 65.4 & \textcolor{blue}{72.9} & \textcolor{blue}{73.1} & \textcolor{blue}{75.7} & 73.8 & - & - & - & - & - & - & -& - & - \\

\midrule
\rowcolor{gray!20}
\textbf{VideoChat-M1 (37B)}                           & \textbf{82.3 } & \textbf{ 84.2}  &   \textbf{76.7}&   \textbf{83.2} & \textbf{83.4}  & \textbf{5.92} &  \textbf{60.5} &   \textbf{80.0} & \textbf{60.4}   &    \textbf{88.3}   &   \textbf{70.8}   &    \textbf{66.7}   & \textbf{71.9} &  \textbf{67.7} \\

\bottomrule
\end{tabular}
}
\vspace{-0.8em}

\caption{Algorithm Comparison. Our VideoChat-M1 results are bolded, and the best results of each group of methods are marked in blue.}

\vspace{-1.5em}

\label{tab:sota_table}
\end{table*}

  \textbf{Result Reward $R_{res}$:}
  After completing our CPP pipeline,
  all agents generate their respective answers.
  We assign a positive reward for correct final answers and a negative penalty for incorrect ones. 

  \textbf{Format Reward $R_{format}$:} To ensure procedural reliability and system compatibility, $R_{format}$ incentivizes syntactically correct actions. It grants rewards for well-formed, executable outputs (e.g., parsable plans, valid tool calls) and imposes penalties for format-related errors.
  
  \textbf{Collaboration Reward $R_{col}$:} 
   To encourage effective agent collaboration beyond the final outcome, we evaluate the intermediate planning process recorded in each agent's memory buffer. We leverage GPT-4o as an external evaluator to assess the holistic quality of the trajectory, including plan feasibility, tool call appropriateness, and step management soundness. To mitigate the inherent stochasticity of LLM-based scoring and ensure a stable learning signal, we constrain the evaluator's output to a binary reward: 1 for coherent trajectories and 0 otherwise (see Appendix A.3 for the prompt). Furthermore, to explicitly promote concise strategies and prevent reward hacking through lengthy planning, we apply a strong penalty to trajectories exceeding five tool calls. Since each agent's memory is influenced by team communication, this reward mechanism incentivizes the entire group to cooperate on developing coherent and efficient policy plans.

After specifying the reward formulation,
we train the agent team using Group Relative Policy Optimization (GRPO)~\cite{deepseek-r1}.
Specifically,
each agent generates $K$ policy plans, producing $K$ candidate final answer,
i.e.,
$o = \{o_1, o_2, ..., o_K\}$.
The advantage $A_R(o_k)$ of each output $o_k \in o$ is then computed
by standardizing its reward against the statistics of all outputs in the group:
\begin{equation}
\setlength\abovedisplayskip{1pt}
\setlength\belowdisplayskip{1pt}
    A_R^{(k)}= 
    \frac{R(o_k)-\mathrm{mean}(\{R(o_1),...,R(o_K)\})}{\mathrm{std}(\{R(o_1),...,R(o_K)\})}
\label{eq:ro}
\end{equation}
Finally, we optimize the model parameters of each agent by maximizing the GRPO objective function:
\begin{equation}
\setlength\abovedisplayskip{1pt}
\setlength\belowdisplayskip{1pt}
    \max_{\pi_\theta} \mathbb{E}_{o\sim \pi_{\theta_{\mathrm{old}}}} \Big[
         \sum_{k=1}^{K} \frac{\pi_\theta (o_k)}{\pi_{\theta_{\mathrm{old}}}(o_k)} \cdot A_R^{(k)}  - \beta\, \mathrm{D}_\mathrm{KL}\Big(\pi_\theta \,\Vert\, \pi_\mathrm{ref}\Big)
    \Big]
    \nonumber
\label{eq:grpo}
\end{equation}
The GRPO objective function balances two components: a reward-seeking term that encourages high-scoring responses, and a KL-divergence penalty that regularizes the policy. This penalty, weighted by the coefficient $\beta$, constrains the optimized policy $\pi_\theta$ of agent $\mathcal{G}_i$ to remain close to the reference policy $\pi_{ref}$, ensuring training stability.
This MARL encourages each agent to refine its policies and collaborate flexibly to answer user queries about the video.

\section{Experiments}
\label{sec:exp}

\noindent\textbf{Datasets.}
We conducted evaluations on 8 video understanding benchmarks described as follows: 
MLVU-Dev~\cite{zhou2024mlvu} includes 2,174 multiple-choice questions and 417 open-ended questions with videos averaging 930 seconds. LongVideoBench~\cite{wu2024longvideobench} provides 1,337 multi-domain QA pairs and videos averaging 473 seconds. 
VSI-Bench~\cite{yang2025thinking} focuses on spatial-temporal reasoning with 2,500 QA pairs that require fine-grained inference of object interactions and temporal causality. VideoMME~\cite{videomme} offers 900 videos (11s-1h) with 2,700 QA pairs,
and MMR-V~\cite{zhu2025mmr} consists of 1257 QA pairs in test set, emphasizing cross-modal and multi-step reasoning.
VideoMMMU~\cite{hu2025video} provides 900 QA pairs for video reasoning,
while Video-Holmes~\cite{cheng2025video} comprises 270 suspense films and 1,837 QA pairs to evaluate complex reasoning via cross-temporal visual clue integration. 
Charades-STA~\cite{gao2017tall} is a large-scale dataset for evaluating temporal grounding tasks with 4233 QA pairs.

\noindent\textbf{Metrics.}
In Tab~\ref{tab:sota_table}, accuracy is adopted as the primary evaluation metric across all benchmarks.
For MMR-V, the CoT column reflects multi-step and compositional reasoning ability by measuring performance under Chain-of-Thought reasoning.
In MLVU, M-avg and G-avg stand for the arithmetic and geometric mean accuracy across multiple sub-tasks, respectively.
For Video-MME, S/M/L correspond to short-, medium-, and long-duration videos.
In VSI-Bench, Dist, Dir, and Order denote reasoning categories for spatial distance, direction, route and temporal order, with Avg representing the overall accuracy.
For Charades, we report the mean Intersection-over-Union between predicted and ground-truth temporal segments.

\noindent\textbf{Implementation Details.}
For our training and testing, we utilized a setup of eight A100 80G GPUs. The learning rates for SFT and MARL was set to 1e-6 and 1e-7.
We performed one epoch of SFT with batch size 32 for each agent on our collected dataset. The best performance was achieved with 200 steps of Multi-Agent Reinforcement Learning (MARL) with 4 rollouts and 8 batch size.
To enhance the generalization of collaboration and avoid co-adaptation, we apply agent dropout. At each training step, a random DAG is sampled from the fully connected agent graph to define the communication topology. This dynamic structure encourages agents to develop robust and flexible communication strategies.
Agent Teams for each task, visualizations and further details are provided in Appendix A.6.

\begin{table}[t]

    \centering
    \small
    \setlength{\tabcolsep}{1mm}
    \renewcommand{\arraystretch}{0.85}
    \begin{tabular}{l|cccc}
        \toprule
        \textbf{Model} & \textbf{Frames} & \begin{tabular}[c]{@{}c@{}} \textbf{Inference}\\ \textbf{Time}  \end{tabular}  &  \begin{tabular}[c]{@{}c@{}} \textbf{LongVideo}\\ \textbf{Bench}  \end{tabular} & \begin{tabular}[c]{@{}c@{}} \textbf{Video}\\ \textbf{MME}  \end{tabular} \\
        \midrule
        Qwen2-VL-72B\cite{qwen2vl}   & 568 &     90.5s    & 55.6        &  71.2 \\
        
        GPT-4o\cite{gpt4o}      & 384 &     153.6s       & 66.7     &  71.9    \\

        Gemini-1.5-Pro\cite{gemini15}   & 568 &    227.2s    & 64.0  &  75.0    \\
        
        \rowcolor{gray!20}
        \textbf{VideoChat-M1}  & \textbf{69.9}  &  \textbf{19.8s}  &  \textbf{82.3} & \textbf{83.2} \\

        \bottomrule
    \end{tabular}
\vspace{-0.75em}
    
    \caption{Average Frame Number and Inference Latency.}
    \label{tab:efficiency}

\vspace{-1em}

\end{table}

\begin{table}[!]

\small
\centering
\setlength{\tabcolsep}{0.9mm}
\renewcommand{\arraystretch}{0.5}

\begin{tabular}{c|cccc|cc}
\toprule
\begin{tabular}[c]{@{}c@{}}\textbf{Num}\\\textbf{Agents}  \end{tabular}  & \begin{tabular}[c]{@{}c@{}} \textbf{Qwen}\\ \textbf{2.5-3B}  \end{tabular}  & \begin{tabular}[c]{@{}c@{}} \textbf{Qwen}\\ \textbf{2.5-7B}  \end{tabular}  & \begin{tabular}[c]{@{}c@{}} \textbf{Qwen}\\ \textbf{3-4B}  \end{tabular} & \begin{tabular}[c]{@{}c@{}} \textbf{Qwen}\\ \textbf{3-8B}  \end{tabular}  & \begin{tabular}[c]{@{}c@{}} \textbf{Video}\\ \textbf{Holmes}  \end{tabular} & \begin{tabular}[c]{@{}c@{}} \textbf{LongVideo}\\ \textbf{Bench}  \end{tabular}\\
    
\midrule

 \multirow{4}{*}{1} & \Checkmark  &       &              &    & 27.8  & 59.2 \\
&  &  \Checkmark     &              &    & 29.9  & 61.1 \\
&  &    &    \Checkmark            &    & 28.9  & 60.4\\ 

 &  &      &   & \Checkmark              &  31.2 & 61.9 \\ \midrule
 
\multirow{6}{*}{2} 

&\Checkmark  &  \Checkmark  &       &     & 41.8 & 66.8 \\ 
&\Checkmark  &   &  \Checkmark      &     & 41.4 & 65.9 \\ 
&\Checkmark  &   &       &   \Checkmark  & 42.3 & 67.2 \\ 
& &  \Checkmark  &   \Checkmark      &     & 42.4 & 67.1  \\  
& &  \Checkmark  &        &   \Checkmark    & 43.5 & 67.9 \\  
& &   &   \Checkmark      &   \Checkmark    &  42.9&  67.2\\  \midrule

\multirow{4}{*}{3}

&\Checkmark   &  \Checkmark   &   \Checkmark      &     &54.8 & 77.2   \\ 
&\Checkmark &  \Checkmark   &        &   \Checkmark    & 55.3 & 78.2 \\ 
&\Checkmark  &   &   \Checkmark      &   \Checkmark    & 55.1 & 77.8 \\ 
& &  \Checkmark &   \Checkmark      &   \Checkmark    & 55.9 & 78.9 \\ \midrule

\rowcolor{gray!20}
4 &\Checkmark   &  \Checkmark   &   \Checkmark      &   \Checkmark   & \textbf{60.5} & \textbf{82.3} \\ 

\bottomrule
\end{tabular}

\vspace{-0.75em}
\caption{Effects of Agent Group Composition and Scale.}
\label{tab:diff_agent}
\vspace{-2em}
\end{table}

\subsection{Comparison with SOTA}
\noindent\textbf{Performance Comparison.}
As shown in Tab~\ref{tab:sota_table}, on models under the 80B scale, we achieved SOTA on 8 datasets. Our VideoChat-M1 approach achieves SOTA on LongVideoBench, outperforming Gemini 2.5 Pro and GPT-4o by 3.6\% and 15.6\%, respectively. It also achieves SOTA performance on the Video-Holmes and MMR-V benchmarks with gains of 14.8\% and 14.3\%. In specialized tasks, our model also achieves the best performance, with a 2.4\% improvement on the VSIBench spatial intelligence task and a 1.8\% lead on the Charades Temporal Grounding task. 
Notably, our efficient 37B model delivers performance comparable to much larger models (such as the Qwen3-VL-235B, Gemini 2.5 pro) on the Video-MME, MLVU, and VideoMMMU benchmarks. Our method uses a CPP mechanism for task decomposition and Multi-Agent Reinforcement Learning (MARL) to enhance cooperation and communication, boosting the group's collective effectiveness.

\begin{table}[t]

\small
\centering
\setlength{\tabcolsep}{0.9mm}
\renewcommand{\arraystretch}{0.8}

\begin{tabular}{c|cccc|cc}
\toprule
\begin{tabular}[c]{@{}c@{}}\textbf{Num}\\\textbf{Agents}  \end{tabular}  & \begin{tabular}[c]{@{}c@{}} \textbf{Qwen}\\ \textbf{2.5-3B}  \end{tabular}  & \begin{tabular}[c]{@{}c@{}} \textbf{Qwen}\\ \textbf{2.5-7B}  \end{tabular}  & \begin{tabular}[c]{@{}c@{}} \textbf{Qwen}\\ \textbf{3-4B}  \end{tabular} & \begin{tabular}[c]{@{}c@{}} \textbf{Qwen}\\ \textbf{3-8B}  \end{tabular}  & \begin{tabular}[c]{@{}c@{}} \textbf{Video}\\ \textbf{Holmes}  \end{tabular} & \begin{tabular}[c]{@{}c@{}} \textbf{LongVideo}\\ \textbf{Bench}  \end{tabular}\\
    
\midrule

\multirow{7}{*}{4}
&$\times$ 2   &     & $\times$ 2      &      & 55.9 & 79.3 \\ 
\rowcolor{gray!20}
&  & $\times$ 2   &        &   $\times$ 2  &  \textbf{58.8} & \textbf{80.9} \\ 
& $\times$ 2 & $\times$ 2 &         &      & 56.0 & 79.4 \\ 
&   &  &    $\times$ 2   &  $\times$ 2    & 56.2 &  79.7\\ 
&   & $\times$ 1  &         &  $\times$ 3   & 57.4 & 80.5 \\ 
&   &   &  $\times$ 1   &$\times$ 3     & 57.2   & 80.1 \\ 
&   &   &      &$\times$ 4     &  55.8  & 79.2 \\ 
\bottomrule
\end{tabular}

\vspace{-0.75em}
\caption{Impact of Architectural Diversity in the 4-Agent Group.}
\label{tab:four_agent}
\vspace{-1em}
\end{table}

\begin{table}[!t]
\small
\centering
\setlength{\tabcolsep}{0.9mm}
\renewcommand{\arraystretch}{0.9}

\begin{tabular}{l|cc}
\toprule
 \textbf{Agent Group } & \begin{tabular}[c]{@{}c@{}} \textbf{Video}\\ \textbf{Holmes}  \end{tabular} & \begin{tabular}[c]{@{}c@{}} \textbf{LongVideo}\\ \textbf{Bench}  \end{tabular}\\ \midrule
 1$\times$ GPT-4o~\cite{gpt4o} + 1$\times$ DeepseekR1~\cite{deepseek-r1} &  51.6 & 71.8  \\
2$\times$ GPT-4o~\cite{gpt4o} + 2$\times$ DeepseekR1~\cite{deepseek-r1} &  56.2 &  75.9 \\
4$\times$ Deepseek-R1~\cite{deepseek-r1}&  51.8 & 71.4 \\ 
4$\times$ GPT-4o~\cite{gpt4o} & 52.7  & 72.9  \\
\rowcolor{gray!20}
    
VideoChat-M1 &  \textbf{60.5} & \textbf{82.3}  \\

\bottomrule
\end{tabular}

\vspace{-0.75em}
\caption{Comparison with Foundation LLM Agent Groups.}
\label{tab:close_source}
\vspace{-1em}
\end{table}

\begin{table}[!t]
\centering
\small
\setlength{\tabcolsep}{1mm}
\renewcommand{\arraystretch}{0.45}
\begin{tabular}{cccc|cc}
\toprule
\textbf{$R_{format}$} & \textbf{$R_{col}$} & \textbf{$R_{res}$} & \begin{tabular}[c]{@{}c@{}} \textbf{Agent}\\ \textbf{Dropout}  \end{tabular}  & \begin{tabular}[c]{@{}c@{}} \textbf{Video}\\ \textbf{Holmes}  \end{tabular} & \begin{tabular}[c]{@{}c@{}} \textbf{LongVideo}\\ \textbf{Bench}  \end{tabular} \\
\midrule                  
\Checkmark & \Checkmark & \XSolidBrush &\Checkmark & 32.4 & 63.8  \\
\Checkmark & \XSolidBrush& \Checkmark &\Checkmark & 59.4 & 81.1  \\
\XSolidBrush& \Checkmark & \Checkmark &\Checkmark & 60.2  &  82.0 \\
\Checkmark & \Checkmark & \Checkmark & \XSolidBrush & 58.5 & 79.9 \\
\rowcolor{gray!20}
\Checkmark & \Checkmark & \Checkmark & \Checkmark & \textbf{60.5} & \textbf{82.3} \\
\bottomrule
\end{tabular}
\vspace{-0.75em}
\caption{Ablation on Components of MARL.}
\vspace{-1em}
\label{tab:finetune}
\end{table}

\begin{table*}[!h]
  \centering

  \begin{minipage}[t]{0.32\textwidth}
    \centering
    \small
    \setlength{\tabcolsep}{1.2mm}
    \renewcommand{\arraystretch}{0.5}  
    \begin{tabular}{cc|cc}
      \toprule
      \textbf{SFT} & \textbf{MARL} &
      \begin{tabular}[c]{@{}c@{}}\textbf{Video}\\ \textbf{Holmes}\end{tabular} &
      \begin{tabular}[c]{@{}c@{}}\textbf{LongVideo}\\ \textbf{Bench}\end{tabular} \\
      \midrule
      \XSolidBrush & \XSolidBrush & 52.1 & 69.3 \\
      \Checkmark   & \XSolidBrush & 55.2 & 75.9  \\
      \XSolidBrush & \Checkmark   & 57.9 & 80.2 \\
      \rowcolor{gray!20}
      \Checkmark   & \Checkmark  & \textbf{60.5} & \textbf{82.3} \\
      \bottomrule
    \end{tabular}
    \vspace{-0.5em}
    
    \caption{Ablation on SFT and RFT.}
    \vspace{-1em}
    \label{tab:sftrft}
  \end{minipage}
  \hfill
  \begin{minipage}[t]{0.32\textwidth}
    \centering
    \small
    \setlength{\tabcolsep}{1.2mm}
    \renewcommand{\arraystretch}{0.95}
    \begin{tabular}{cc|cc}
      \toprule
      \begin{tabular}[c]{@{}c@{}}\textbf{Full}\\ \textbf{Finetune}\end{tabular} &
      \textbf{LoRA} &
      \begin{tabular}[c]{@{}c@{}}\textbf{Video}\\ \textbf{Holmes}\end{tabular} &
      \begin{tabular}[c]{@{}c@{}}\textbf{LongVideo}\\ \textbf{Bench}\end{tabular} \\
      \midrule
      
      \XSolidBrush & \XSolidBrush & 55.2 & 75.9 \\
      \XSolidBrush & \Checkmark  & 59.4& 81.2 \\
      
      \rowcolor{gray!20}
      \Checkmark   & \XSolidBrush & \textbf{60.5} & \textbf{82.3} \\
      \bottomrule
    \end{tabular}
    \vspace{-0.5em}
    \caption{Different tuning methods.}
    \vspace{-1em}
    \label{tab:lora_ft}
  \end{minipage}
  \hfill
  \begin{minipage}[t]{0.32\textwidth}
    \centering
    \small
    \setlength{\tabcolsep}{0.9mm}
    \begin{tabular}{c|cc}
      \toprule
      \begin{tabular}[c]{@{}c@{}}\textbf{Decision}\\ \textbf{Methods}\end{tabular} &
      \begin{tabular}[c]{@{}c@{}}\textbf{Video}\\ \textbf{Holmes}\end{tabular} &
      \begin{tabular}[c]{@{}c@{}}\textbf{LongVideo}\\ \textbf{Bench}\end{tabular} \\
      \midrule
      Best Score       &  59.9 &  81.2\\
      Decide by Agent  & 60.2 & 81.6\\
      \rowcolor{gray!20}
      \textbf{Vote}    & \textbf{60.5} & \textbf{82.3}  \\
      \bottomrule
    \end{tabular}
    \vspace{-0.5em}
    
    \caption{Different discussion mechanisms.}
    \label{tab:final_dis}
  \end{minipage}

  \vspace{-1.5em}

  \label{tab:ablations_all}
\end{table*}

\begin{figure}[t]
    \centering
    \includegraphics[width=0.45\textwidth]{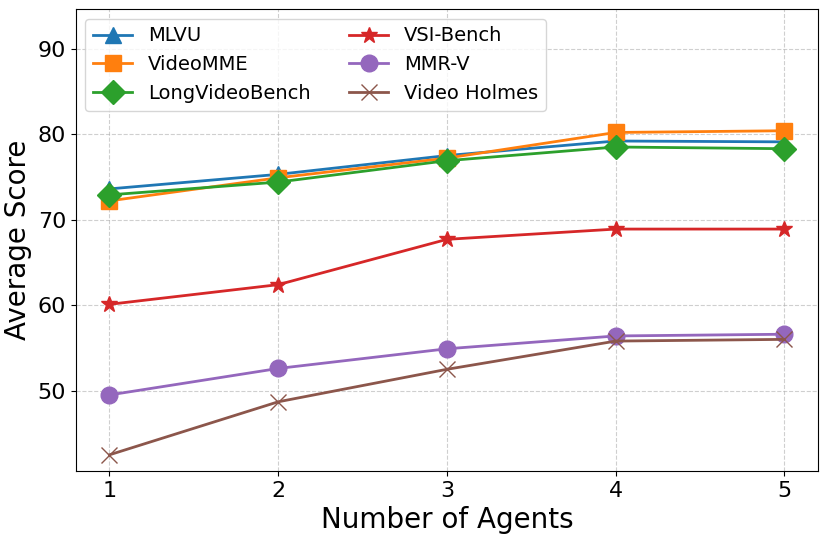}
    \vspace{-1em}
    \caption{Effects of the Number of Homogeneous Agents.}
    \vspace{-1.5em}
    
    \label{fig:num_agents}
\end{figure}

\noindent\textbf{Efficiency Comparison.}
From Tab~\ref{tab:efficiency}, we observe that VideoChat-M1 uses only 69.9 frames per video, accounting for 12.3\%$\sim$18.2\% of other models. Meanwhile, its average inference time is 19.8s, which is merely 8.7\%$\sim$21.9\% of the baselines. Notably, despite the significantly reduced computational cost, VideoChat-M1 still achieves top scores on LongVideoBench (82.3\%) and VideoMME (83.2\%), highlighting its superior efficiency–performance trade-off.

\subsection{Ablation Studies}

\noindent\textbf{Effects of the Number of Homogeneous Agents.}
To investigate how the number of agents affects performance, we conducted experiments using the best-performing Qwen3-8B architecture.
As shown in Fig~\ref{fig:num_agents}, performance improves steadily as the number of agents increases from one to four. However, further increasing the number beyond four leads to performance saturation, with negligible additional gains.

\noindent\textbf{Effects of Agent Group Composition and Scale.} To investigate the influence of agent group composition and scale, we conducted experiments with diverse configurations (Tab~\ref{tab:diff_agent}). Our findings reveal two key trends: first, performance consistently improves as the total number of agents increases. Second, for groups of the same size, those with a larger parameter capacity achieve superior results.

\noindent\textbf{{Impact of Architectural Diversity within Agent Groups.}}
As shown in Tab~\ref{tab:four_agent}, we further explore the impact of architectural diversity within a 4-agent setup. We consider configurations where some agents share identical architectures against groups composed of entirely distinct agents. Experimental results indicate that structural redundancy among agents reduces discussion diversity, leading to diminished collaborative gains compared to fully heterogeneous groups. Thus, even with a smaller overall parameter count, our CPP paradigm, when applied to diverse agent groups, enables greater performance improvements than cooperation among homogeneous agents.

\noindent\textbf{Comparison with Foundation LLM Agent Groups.}
Tab~\ref{tab:close_source} reports the performance of untrained close-sourced foundation LLM teams following the same CPP protocol. Two GPT-4o and two DeepSeek-R1 agents achieve 56.2 and 75.9, respectively, while homogeneous teams of four GPT-4o or four DeepSeek-R1 remain below 53 and 73. VideoChat-M1, trained with MARL, outperforms them by at least 4.3 and 6.4 points. The gap verifies that collaborative fine-tuning injects task-specific coordination patterns that even stronger proprietary models fail to discover via zero-shot reasoning alone.

\noindent\textbf{Ablation Study on Key Components of MARL.}
Tab~\ref{tab:finetune} evaluates the individual contributions of the reward components and agent dropout. Removing the process reward drops performance by one point on both benchmarks, while omitting the format reward causes a similar degradation. Disabling agent dropout incurs a larger penalty of two points, indicating that dynamic topology is the most critical regularizer. The full configuration yields the highest scores, confirming that dense process feedback and stochastic communication are both necessary for optimal collaborative policy learning.

\noindent\textbf{Ablation on SFT and RFT.}
Tab~\ref{tab:sftrft} disentangles the contributions of supervised fine-tuning (SFT) and collaborative reinforcement learning (RFT). The foundation model without either stage achieves 52.1 and 69.3. Applying SFT alone boosts scores to 55.2 and 75.9, while RFT alone reaches 57.9 and 80.2. The full pipeline, which first establishes reliable planning priors through SFT and then refines inter-agent coordination via RFT, attains peak performance of 60.5 and 82.3. These additive gains confirm that principled initialization (from SFT) and emergent collaboration (from RFT) are both indispensable for maximum performance.

\noindent\textbf{Ablation on Finetuning Strategies.}
To validate the necessity of full-parameter finetuning, we compare its performance with LoRA that only updates about 2\% of the training parameters. As shown in Tab~\ref{tab:lora_ft}, while full-parameter finetuning yields slightly superior performance, the marginal gap confirms that our collaborative policy can be successfully implanted by tuning just this small subset of parameters. This highlights LoRA as a lightweight deployment option without significant accuracy loss.

\noindent\textbf{Ablation on Discussion Mechanisms.}
Tab~\ref{tab:final_dis} evaluates three different discussion mechanisms for aggregating individual agent conclusions. 
The first (``Best Score") involves each agent scoring its own and others’ responses, with the highest-scoring result selected (59.9/81.2). The second (``Decide by Agent") directly adopts the output of Qwen3-8B (chosen for its strongest performance), yielding 60.2/81.6. The third (``Vote") selects the majority-endorsed answer, which further elevates performance to 60.5/82.3. This confirms that the diversity generated by independent agent planning is best leveraged through lightweight majority consensus, outperforming score-based or authority-based selection strategies.

\section{Conclusion}

We introduce VideoChat-M1, a novel multi-agent framework for adaptive tool invocation in video understanding. Built on a Collaborative Policy Planning (CPP) paradigm and trained with a streamlined Multi-Agent Reinforcement Learning (MARL) approach, the framework dynamically discovers critical clues to achieve robust video reasoning. VideoChat-M1 achieves SOTA performance across eight benchmarks on four mainstream video tasks: long-form video QA, video reasoning, spatial intelligence, and temporal grounding. 
To the best of our knowledge, this is the first multi-agent policy learning framework for tackling complex video understanding tasks, contributing to the development of more adaptive and intelligent video understanding.

\label{sec:conclusion}

\section{Acknowledgements}
{
\raggedright
This work was supported by Guangdong Science and Technology Program (Grant No. 2024TQ08X365) \par
}
{
    \small
    \bibliographystyle{ieeenat_fullname}
    \bibliography{main}
}

\clearpage
\setcounter{page}{1}
\maketitlesupplementary

\section*{A.1 Collected Dataset}

To equip VideoChat-M1 with strong generalization across diverse video understanding scenarios, we assemble a comprehensive collection of datasets spanning multiple task types, including temporal grounding, long-video question answering, Spatial Intelligence analysis, and video reasoning. These datasets originate from widely used benchmarks and cover a broad spectrum of video durations, scenes, and annotation forms. The diversity of tasks and data sources empowers VideoChat-M1 to learn from heterogeneous supervision signals, enhancing its capabilities in perceiving, retrieving, and reasoning over long and complex videos.  
Table~\ref{tab:dataset} summarizes the instance numbers and average video durations of all datasets used in our training pipeline. 
In total, the dataset collection comprises 102,911 instances with an overall average video duration of 194.6 seconds, laying a solid data foundation for training VideoChat-M1 on four mainstream video tasks.

\begin{table}[t]
\centering
\small 
\setlength{\tabcolsep}{1mm} 

\begin{tabular}{llrr} 
\toprule
\textbf{Type} & \textbf{Dataset} & \makecell{\textbf{Instance} \\ \textbf{Num}} & \makecell{\textbf{Avg Video} \\ \textbf{Length (s)}} \\

\midrule

\multirow{3}{*}{Temporal Grounding} & FineAction & 5067 & 43.64 \\
 & QVHighlights & 13790 & 28.36 \\
 & HiREST & 3617 & 282.45 \\
\midrule

\multirow{4}{*}{Long Video QA} & ActivityNet-QA & 16642 & 621.45 \\
 & LongViTU & 16453 & 268.46 \\
  & MMBench & 1673 & 97.51 \\
 & MovieChat & 808 & 457.65 \\
  & Neptune & 5281 & 149.25 \\
 
\midrule

\multirow{2}{*}{Spatial Intelligence} & HoursVideo & 831 & 568.16 \\
 & SpaceR & 12643 & 10.65 \\
\midrule

\multirow{3}{*}{Video Reasoning} 
& Video-R1 & 15123 & 68.56 \\
 & VideoEspresso & 9432 & 56.12 \\
  & \begin{tabular}[c]{@{}c@{}}Video Holmes\\ (Training Set)\end{tabular} & 1551 & 91.16 \\
\midrule

\textbf{Total} & \textbf{--} & \textbf{102911} & \textbf{194.6} \\ 
\bottomrule
\end{tabular}
\caption{Instance numbers of different datasets for VideoChat-M1 training.}
\label{tab:dataset}
\vspace{-1em}
\end{table}

\section*{A.2 Memory Buffer and Tool Use}

\begin{figure*}
    \centering
    \includegraphics[width=\linewidth]{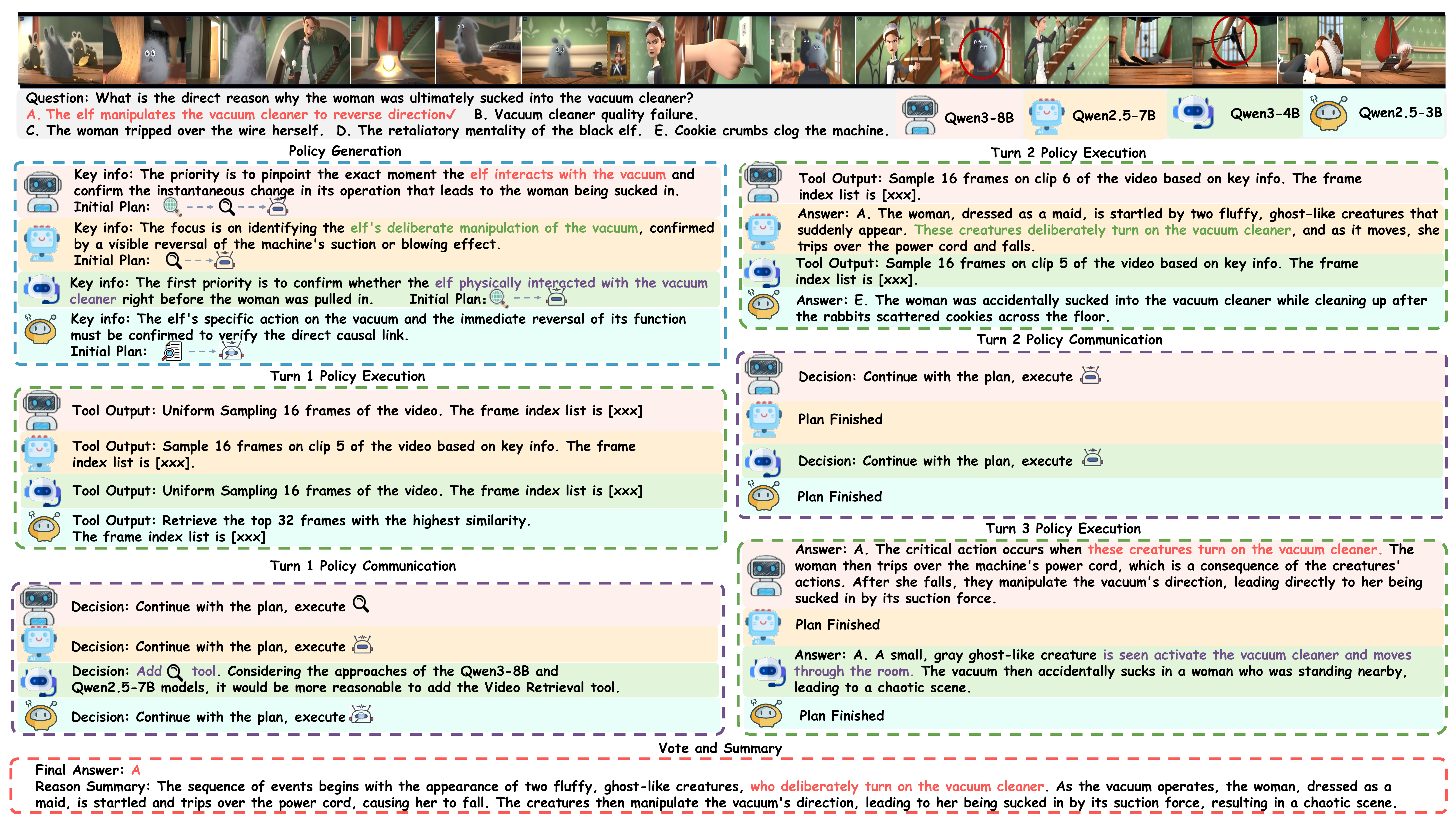}
    \vspace{-2em}
    
    \caption{Visualization of VideoChat-M1 at each step of the CPP process.
    }
    \label{fig:cpp_step_vis}
\end{figure*}

We implement the memory buffer as a key-value pair structure, in which keys denote the agents' names and values store the structured information illustrated in Fig~\ref{fig:cpp_step_vis}. We take the memory buffer of Qwen3-8B as an example.

\begin{tcolorbox}[
    colback=white, 
    colframe=black, 
    sharp corners, 
    boxrule=0.8pt, 
    breakable,
    title={Memory Buffer}, 
    fonttitle=\bfseries\large,            
    coltitle=white,                       
]

\textbf{Agent Name:} Qwen3-8B

The initial plan is: \texttt{<Global Sampling>, <Video Retrieval>, <Rough Browser>}

Key info: The priority is to pinpoint the exact moment the elf interacts with the vacuum and confirm the instantaneous change in its operation that leads to the woman being sucked in.

\textbf{Turn 1}: \\
Executing Tool: \texttt{Global Sampling}

Tool Output: Uniform Sampling 16 frames of the video. The frame index list is [xxx]

Communication Decision: Continue with the plan, execute \texttt{Video Retrieval}

\textbf{Turn 2}: \\
Executing Tool: \texttt{Video Retrieval}

Tool Output: Sample 16 frames on clip 6 of the video based on key info. The frame index list is [xxx]. 

Communication Decision: Continue with the plan, execute \texttt{Rough Browser}

\textbf{Turn 3}: \\
Executing Tool: \texttt{Rough Browser}

\textbf{Answer:} A

\textbf{Summary:} The critical action occurs when these creatures turn on the vacuum cleaner. The woman then trips over the machine's power cord, which is a consequence of the creatures' actions. After she falls, they manipulate the vacuum's direction, leading directly to her being sucked in by its suction force.

\textbf{Plan Finished}

\end{tcolorbox}

To enable our multi-agent framework to tackle a diverse array of video understanding tasks, we provide each agent with access to a comprehensive and specialized toolkit $\mathcal{T}$. These tools facilitate efficient information extraction, spanning coarse-grained retrieval to fine-grained perceptual analysis. The tools available are as follows:

\begin{itemize}
    \item \textbf{Global Sampling:} For queries requiring a holistic understanding, this tool uniformly samples frames across the entire video duration.     
    \item \textbf{Video Retrieval:} This tool first divides the video into six equal-length clips. It then employs the ASP-CLIP~\cite{asp} model to score the semantic similarity between each clip and the user query, returning the highest-scoring clip for further analysis.   
    \item \textbf{Time Stamp Retrieval:} When a precise moment is referenced, this tool extracts a one-minute video segment centered at the specified timestamp.     
    \item \textbf{Image Retrieval:} For the image retrieval stage, we uniformly sample frames from the source video at a rate of 2 frames per second (fps). We then employ the pre-trained CLIP model to compute the similarity score between the textual prompt and each sampled frame, ultimately selecting the top 16 (or 32) frames with the highest similarity.
    
    \item \textbf{Rough Browser:} This tool provides a rapid overview by processing a sparse set of 16 selected frames with a Multimodal Large Language Model (MLLM), such as Qwen2.5-VL-7B~\cite{Qwen2.5-VL}.          
    \item \textbf{Fine Browser: }For detailed analysis where a deeper look is necessary, this tool leverages the same MLLM to process a denser sequence of 32 frames extracted from a targeted video clip.           \item \textbf{Spatial Tool:} To address spatial reasoning queries, this tool employs the InternVL-3.5-8B~\cite{wang2025internvl35} model to analyze 16 frames, which are either uniformly sampled or sourced from a retrieved clip.
        
    \item \textbf{Grounding Tool:} This specialized tool is designed for temporal grounding tasks and utilizes the Eagle2.5-7B~\cite{eagle_2.5} model to process the video and identify relevant time segments.
    
\end{itemize}

\section*{A.3 Prompt}

In this section, we detail the prompts employed in each step of our proposed method.

\begin{tcolorbox}[
    colback=white, 
    colframe=black, 
    sharp corners, 
    boxrule=0.8pt, 
    breakable,
    title={Prompt for Policy Generation}, 
    fonttitle=\bfseries\large,            
    coltitle=white,                    
]

\noindent You are an intelligent video understanding agent. Your task is to analyze a video question and select the optimal combination of tools to answer it accurately.

\vspace{1em}

\noindent \textbf{1. Tool Definitions}

\noindent \textbf{Group A: Frame Selection Tools (Retrieval Phase)}
\begin{itemize}[leftmargin=*, label={$\bullet$}]
    \item \textbf{Uniform Sampling:} A general strategy. Use this only when the question is broad or covers the whole video. It summarizes the overall content without focusing on specific details.
    \item \textbf{Video Retrieval:} The standard semantic search method. Use this to locate the most relevant video clips containing the action, event, or object described in the text query.
    \item \textbf{Time Stamp Retrieval:} Deterministic retrieval. Use this strictly when the question mentions a specific time (e.g., ``at 01:30'').
    \item \textbf{Image Retrieval:} Fine-grained visual matching. Use this to identify specific static scenes, small objects, or person attributes by matching text descriptions to individual frames (top-k selection).
\end{itemize}

\vspace{1em}

\noindent \textbf{Group B: Video Browsing Tools (Reasoning Phase)}
\begin{itemize}[leftmargin=*, label={$\bullet$}]
    \item \textbf{Rough Browser:} Provides a comprehensive yet efficient overview of the selected frames. Sufficient for answering the majority of general video understanding questions.
    \item \textbf{Fine Browser:} High-computation analysis. Use this \textit{only} for cases of extreme ambiguity or when deciphering subtle details (e.g., small text, rapid motions) is critical.
    \item \textbf{Spatial Tool:} Specialized for spatial reasoning benchmarks (e.g., VSIBench). Use this when the question explicitly asks about relative positions, geometry, or spatial arrangements of objects.
    \item \textbf{Grounding Tool:} Specialized for temporal localization (e.g., Charades-STA). Use this strictly for simple, single-scene grounding tasks where the goal is to identify start/end timestamps rather than complex reasoning.
\end{itemize}

\vspace{1em}

\noindent \textbf{2. Recommended Workflow}

\noindent You \textbf{MUST} adhere to the following selection rules:
\begin{enumerate}[itemsep=2pt]
    \item \textbf{Selection Phase:} You must select \textbf{ONE} or more tools from Group A (Frame Selection).
    \item \textbf{Browsing Phase:} You must select \textbf{ONE} or more tools from Group B (Video Browsing).
    \item "Analyze the question and candidate options to determine the key information necessary for the reasoning process. This becomes your \textbf{Key info}.
    
\end{enumerate}

\vspace{1em}

\noindent \textbf{Current Task:} \{task\}\\
\textbf{Question:} \{question\}

\noindent \textbf{3. Output Format \& Examples}

\noindent \textit{Example 1 (General Reasoning):} \\
\textbf{Question:} What does the object being chased by the people refer to? \\
\textbf{Options:} A: Difficulties in life, B: His fully automatic house... 

\vspace{1em}

\textbf{Format:} \\
\#\#key info: the object being chased by the people in the video. \\
\#\#tool use: \texttt{<Video Retrieval>}, \texttt{<Rough Browser>}

\vspace{0.5em}

\noindent \textit{Example 2 (Spatial Intelligence):} \\
\textbf{Question:} If I am standing by the table and facing the bathtub, is the bed to my left, right, or back? \\
\textbf{Options:} A: left, B: right, C: Back

\vspace{1em}

\textbf{Format:} \\
\#\#key info: spatial relation between bathtub and bed. \\
\#\#tool use: \texttt{<Video Retrieval>}, \texttt{<Spatial Tool>}

\end{tcolorbox}

\begin{tcolorbox}[
    colback=white, 
    colframe=black, 
    sharp corners, 
    boxrule=0.8pt, 
    breakable,
    title={Prompt for Policy Communication}, 
    fonttitle=\bfseries\large,            
    coltitle=white,                       
]

\noindent You are a strategic planning assistant. Your sole responsibility is to evaluate the current execution state and determine the immediate next step.

\noindent \textbf{1. CURRENT CONTEXT}

\noindent Review the following execution state carefully:
\begin{itemize}[leftmargin=0.5cm, label={\textbf{-}}]
    \item \textbf{Original Question:} \{question\}
    \item \textbf{Memory buffer:} \{memory\}
    \item \textbf{Other Agents' Output:} \{other agents output\}
    \item \textbf{Remaining Plan:} \{plan\}
\end{itemize}

\noindent \textbf{2. DECISION PROTOCOL}

\noindent You must choose exactly \textbf{ONE} action from the list below based on the logic provided:

\vspace{0.5em}

\noindent \textbf{Option A: The Standard Path}
\begin{itemize}[leftmargin=0.5cm]
    \item \texttt{continue()}: 
    Use this to proceed with the \texttt{\{next tool\}}. 
    \textbf{Rule:} Apply this when peer agents offer no constructive alternatives and the current internal plan remains valid and error-free.
\end{itemize}

\noindent \textbf{Option B: Exception Handling}
\begin{itemize}[leftmargin=0.5cm]
    \item \texttt{add tool(tool name='<name>')}: 
    Use this \textbf{ONLY} if the current plan is logically flawed and requires a new tool (e.g., 'Video Retrieval') to proceed. Analyze the question, candidate options, and the memory of all agents to determine the key information necessary for the reasoning process. This becomes the \textbf{Key info}. 
\end{itemize}

\noindent Output your response strictly in the format below.

\vspace{0.5em}
\textbf{Format:}

\noindent \textbf{Scenario 1: Continuing (Default)}

\#\#tool call: continue()

\noindent \textbf{Scenario 2: Adding a Tool (Correction)}

\#\#tool call: add tool \texttt{<Video Retrieval>} \\
\#\#key info: xx

\end{tcolorbox}

\newpage

\begin{tcolorbox}[
    colback=white, 
    colframe=black, 
    sharp corners, 
    boxrule=0.8pt, 
    breakable,
    title={Prompt for Answering the Question}, 
    fonttitle=\bfseries\large,            
    coltitle=white,                       
]

\textsf{(Use this when the agent's plan is fully executed, but the answer remains unresolved)}\\

You are an intelligent agent responsible for synthesizing a final answer based strictly on the provided internal logs, referred to as \texttt{\{Agent Memory\}}. You must adhere to the following format constraints based on the presence of options.

\medskip
\textbf{Input Context:}
\begin{itemize}[leftmargin=*, itemsep=0pt]
    \item \textbf{Question}: \texttt{\{Question\}}
    \item \textbf{Option}: \texttt{\{Option\}} \textit{(Note: If this field is empty, treat as an open-ended task or temporal grounding task.)}
    \item \textbf{Task}: \texttt{\{Task\}}
    \item \textbf{Agent Memory}: \texttt{\{Agent Memory\}}
\end{itemize}

\medskip
\textbf{Directives:}
\begin{enumerate}[leftmargin=*]
    \item \textbf{Source of Truth:} Your response must be derived solely from the information contained within \texttt{\{Agent Memory\}}. Do not hallucinate or use external knowledge.
    \item \textbf{Multiple Choice Logic:} If \texttt{\{Option\}} is provided (e.g., A, B, C, D), your final output must be \textbf{the single uppercase letter} corresponding to the correct choice and the reason for your answer.
    \item \textbf{Open-Ended Logic:} If \texttt{\{Option\}} is not provided (e.g., Temporal Grounding or open-ended QA), your final output must be a paragraph explaining the reasoning for the answer.
\end{enumerate}

\textbf{Format}:

\#\#Answer: xx \\
\#\#Reason: xx

\end{tcolorbox}

\begin{tcolorbox}[
    colback=white, 
    colframe=black, 
    sharp corners, 
    boxrule=0.8pt, 
    breakable,
    title={Prompt for Reason Summary}, 
    fonttitle=\bfseries\large,            
    coltitle=white,                       
]

\textbf{Input Context:}\\
You have been provided with the reasoning from four distinct AI agents:
\begin{itemize}[leftmargin=*, itemsep=2pt]
    \item \textbf{\{Agent0 name\}}: \{agent 0 reason\}
    \item \textbf{\{Agent1 name\}}: \{agent 1 reason\}
    \item \textbf{\{Agent2 name\}}: \{agent 2 reason\}
    \item \textbf{\{Agent3 name\}}: \{agent 3 reason\}
\end{itemize}

\vspace{1em} 

\textbf{Your Task:}\\
Synthesize and summarize the reasons of each agent into a single, cohesive paragraph. 

\textit{Critically, you must adhere to the following synthesis logic based on the question type:}

\begin{enumerate}[label=\arabic*., leftmargin=*]
    \item \textbf{For Multiple Choice Questions (Options provided):} 
    Identify the final consensus option (or the selected answer). You must ONLY summarize the results and reasoning of the agents that agreed with this final option. Ignore the reasoning of dissenting agents unless it provides critical context for the correct answer.
    
    \item \textbf{For Open-Ended Questions (No options provided):} 
    Synthesize and summarize the reasoning from \textbf{ALL} agents to provide a comprehensive answer. In particular, the summarization should prioritize the consensus among agents, placing greater emphasis on convergent reasoning paths found in similar responses.
\end{enumerate}

The final summary must be concise but accurately reflect the sequence of events and core logic.

\vspace{1em}

\textbf{Format:}

\#\#Final Answer: xx\\
\#\#Reason Summary: xx

\end{tcolorbox}

\begin{tcolorbox}[
    colback=white, 
    colframe=black, 
    sharp corners, 
    boxrule=0.8pt, 
    breakable,
    title={Prompt for Rough Browser}, 
    fonttitle=\bfseries\large,            
    coltitle=white,                       
]

\noindent\textbf{Input Components:} \\
You will be provided with the following:
\begin{itemize}
    \item A sequence of key frames extracted from a video.
    \item Question:\{Question\} and Options \{if have Options or None\}
    \item A key info text that specifies the central theme for the summary. \{Key info\}
    
\end{itemize}

\vspace{1em}

\noindent\textbf{Your Task:} \\
Write a brief summary of the video's content. The summary \textbf{must be centered around} the event, object, or action described in the \textbf{Key info}. The entire summary must be \textbf{no more than 128 tokens}. If you can provide the answer to the question, you can also give the answer.

\vspace{1em}

\noindent\textbf{Output Format:} \\
\#\#Answer: xx \\
\#\#Summary: A single, concise paragraph containing the summary.

\end{tcolorbox}

\begin{tcolorbox}[
    colback=white, 
    colframe=black, 
    sharp corners, 
    boxrule=0.8pt, 
    breakable,
    title={Prompt for Fine Browser}, 
    fonttitle=\bfseries\large,            
    coltitle=white,                       
]

\noindent\textbf{Input Components:} \\
\begin{itemize}
    \item A video clip requiring detailed examination.
    \item Question:\{Question\} and Options \{if have Options or None\}
    \item A key info text that directs the model's focus to the most critical aspect of the video for solving the problem. \texttt{\{key info\}}
\end{itemize}

\vspace{1em}

The model's core task is to generate a detailed summary by analyzing 32 uniformly sampled frames from the video clip. This summary must be thematically centered on the event, object, or action specified in the \textbf{Key info}. This fine-grained analysis is specifically designed to resolve high ambiguity and decipher subtle details (e.g., small text, rapid motions) that are critical for a correct interpretation. If you can provide the answer to the question, you can also give the answer.

\vspace{1em}

\noindent\textbf{Output Format:}  \\
\#\#Answer: xx \\
\#\#Summary: The final summary must be concise yet descriptive, and it \textbf{must not exceed  256 tokens}.

\end{tcolorbox}

\begin{tcolorbox}[
    colback=white, 
    colframe=black, 
    sharp corners, 
    boxrule=0.8pt, 
    breakable,
    title={Prompt for Spatial Tool}, 
    fonttitle=\bfseries\large,            
    coltitle=white,                       
]

\noindent\textbf{Input Components:} \\
\begin{itemize}
    \item A video clip requiring detailed examination.
    \item Question:\{Question\} and Options \{if have Options or None\}
    \item A key info text that directs the model's focus to the specific spatial question that needs to be answered. \texttt{\{key info\}}
\end{itemize}

\vspace{1em}

This tool is specifically invoked for queries concerning the relative positions, geometry, or spatial arrangements of objects, as is common in spatial reasoning benchmarks (e.g., VSIBench). To address these queries, the model's core task is to analyze 32 uniformly sampled frames to build a comprehensive understanding of the scene's spatial layout. It must then generate a descriptive summary that explicitly answers the spatial question posed in the \textbf{Key info} by identifying key objects and precisely describing their positions relative to each other. If you can provide the answer to the question, you can also give the answer.

\vspace{1em}

\noindent\textbf{Output Format:} \\
\#\#Answer: xx \\
\#\#Summary: The final summary must be concise yet descriptive, and it \textbf{must not exceed  256 tokens}.

\end{tcolorbox}

\begin{tcolorbox}[
    colback=white, 
    colframe=black, 
    sharp corners, 
    boxrule=0.8pt, 
    breakable,
    title={Prompt for Grounding Tool}, 
    fonttitle=\bfseries\large,            
    coltitle=white,                      
]

Given a user-provided textual key info prompt and a video, the model must retrieve the precise time segment in the video that directly corresponds to the prompt. Furthermore, the model must generate a concise, natural language justification for its selection.

The textual prompt is: \{\textbf{Key info}\}

\noindent\textbf{Output Format:} 

\#\#Timestamp: [xxs - xxs]

\#\#Reason: xxx

\end{tcolorbox}

\begin{figure*}
    \centering
    \includegraphics[width=\linewidth]{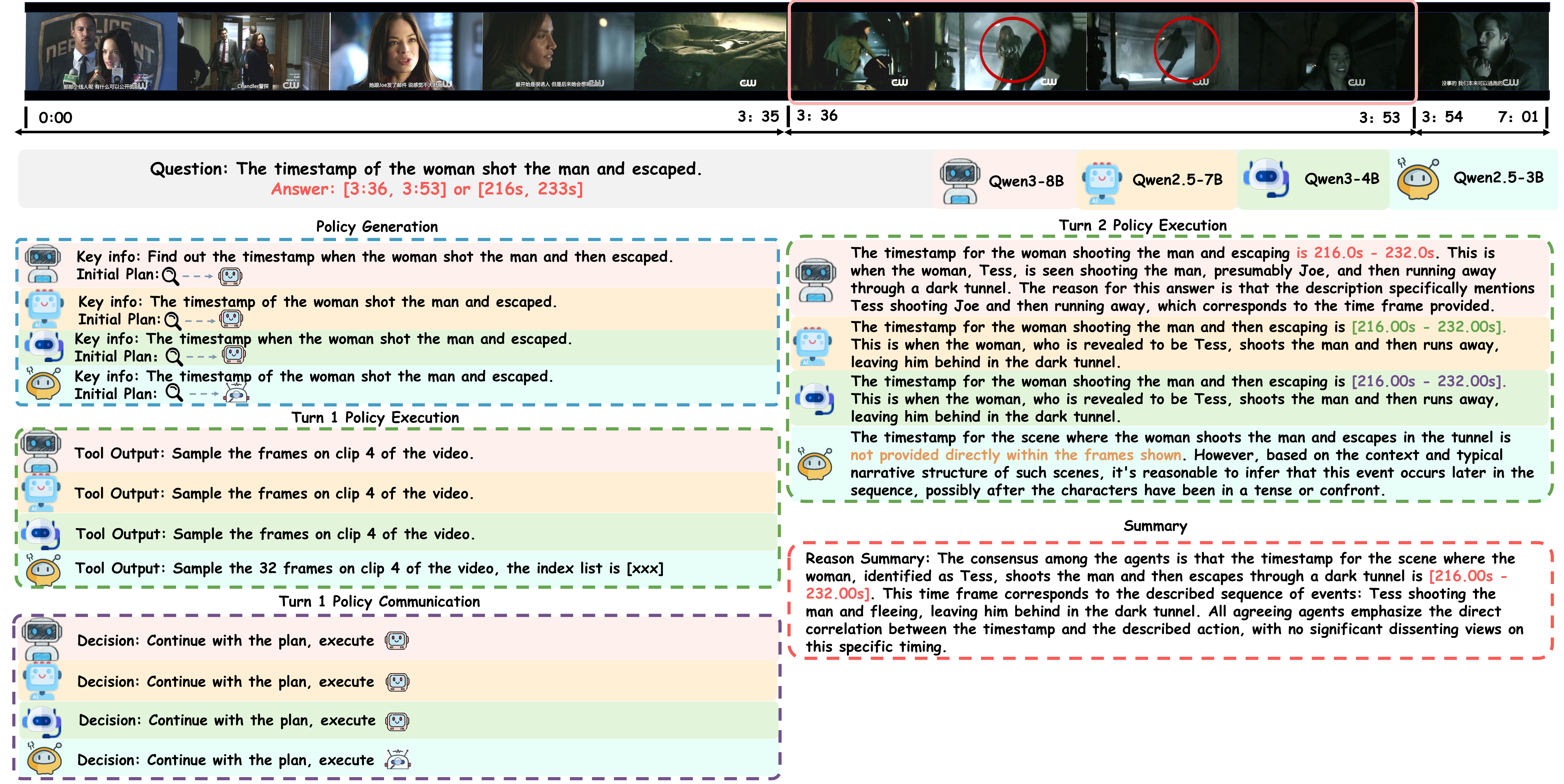}
    \vspace{-2em}
    
    \caption{Visualization of VideoChat-M1 on each step of the CPP process on the temporal grounding task.
    }
    \label{fig:cpp_step_vis_step2}
    \vspace{-1.4em}
\end{figure*}

\begin{figure*}
    \centering
    \includegraphics[width=\linewidth]{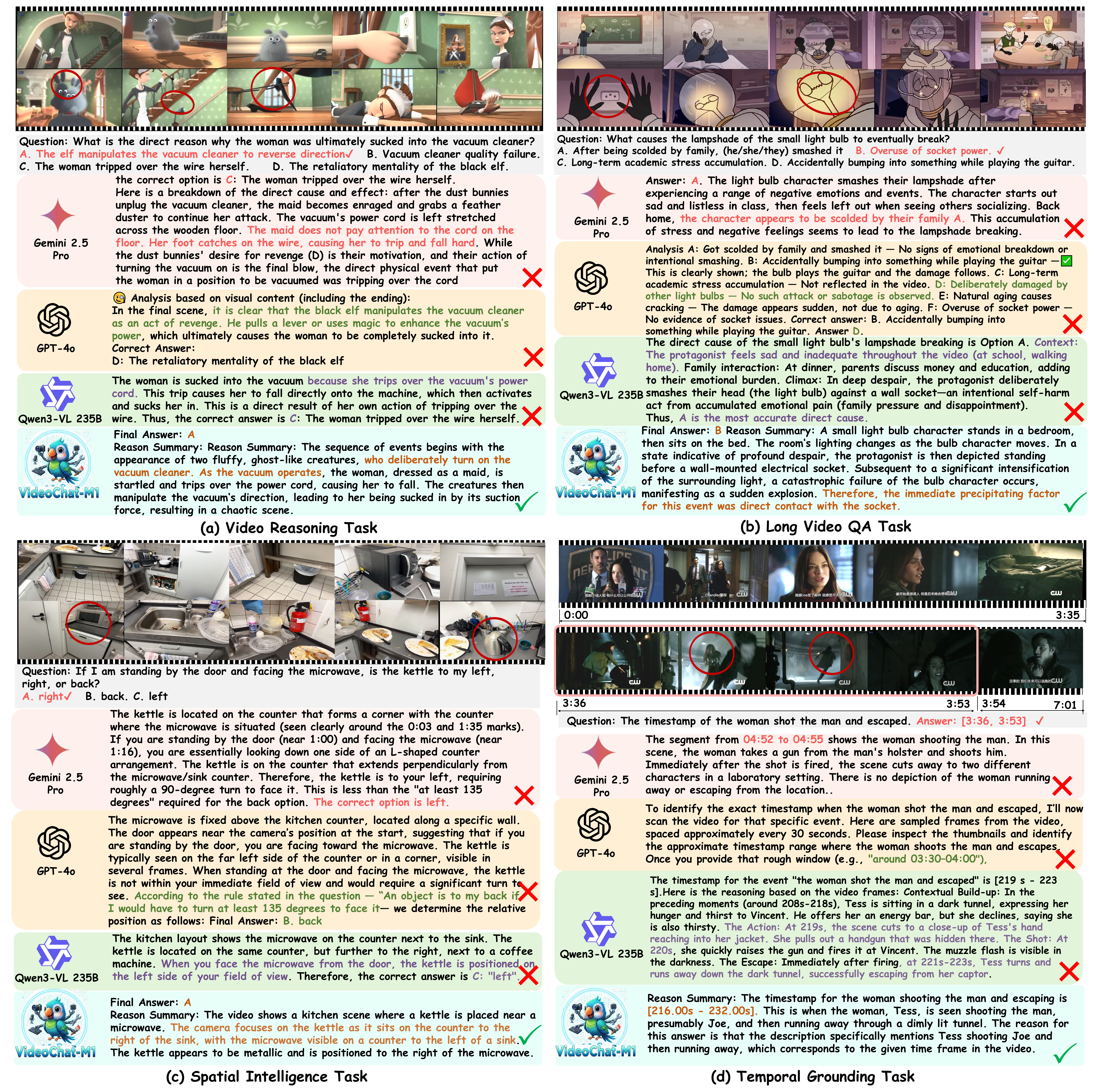}
    \vspace{-2em}
    
    \caption{Visualization of VideoChat-M1 on four mainstream tasks.
    }
    
    \label{fig:qavis}
    \vspace{-1.4em}
\end{figure*}

\section*{A.4 Visualization}
To obtain qualitative insights into our method's mechanics and efficacy, this section presents a two-part visual analysis of VideoChat-M1. First, Fig~\ref{fig:cpp_step_vis} and Fig~\ref{fig:cpp_step_vis_step2} provide a fine-grained visualization of the Collaborative Policy Planning (CPP) process, tracing policy evolution and intermediate reasoning steps to enhance the interpretability of our multi-agent framework. Second, Fig~\ref{fig:qavis} reports a comparative qualitative evaluation, benchmarking the visual outputs of VideoChat-M1 against those of state-of-the-art models across four canonical video understanding tasks. This is intended to empirically validate the performance improvements achieved by our method.

Fig~\ref{fig:cpp_step_vis} details each step of our CPP process and its corresponding output. It demonstrates that our framework can autonomously refine its plans during execution and exhibits a high degree of fault tolerance, enabling the agent group to recover from errors made by individual agents. The final summary is generated by synthesizing the rationales from all agents that voted for the correct answer ('A'), a task facilitated by the Qwen3-8B model.

In Fig~\ref{fig:cpp_step_vis_step2}, we present a step-by-step visualization of our CPP framework applied to the open-ended temporal grounding task. Initially, a video retrieval tool is employed as a coarse-grained filter, significantly constricting the temporal search space to a relevant video clip. Subsequently, our CPP method operates within this narrowed window to perform fine-grained boundary refinement. As demonstrated by the query 'a woman shot the man and escaped,' the retrieval module effectively eliminated irrelevant footage, enabling our model to focus on the semantic context. Consequently, the method precisely localized the target interval, aligning perfectly with the ground truth, although the Qwen2.5-3B agent failed to find the result. 

Fig~\ref{fig:qavis} compares our method with recent Multimodal Large Language Models (MLLMs) across four mainstream tasks. This comparison reveals that existing models frequently rely on superficial cues, miss critical shots, or fail to maintain long-range temporal and spatial consistency, leading to incorrect reasoning. In contrast, VideoChat-M1 reliably identifies causal relations, tracks events over long video durations, infers accurate spatial layouts, and precisely localizes actions in time. These results show that our collaborative, multi-step reasoning framework delivers more accurate, stable, and interpretable video understanding compared to prior approaches.

\begin{figure}
    \centering
    \includegraphics[width=\linewidth]{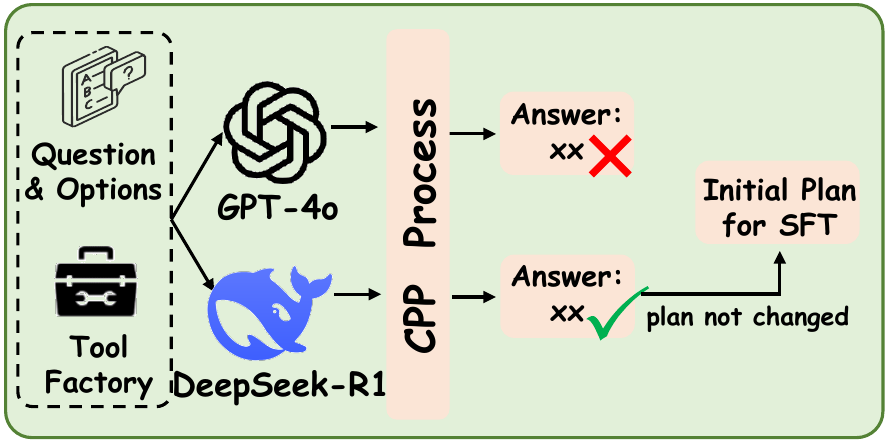}
    \vspace{-1.4em}
    \caption{The process of generating the SFT data.
    }
    \label{fig:sft}
\end{figure}

\section*{A.5 Reinforcement Learning Analysis}

While a formal convergence proof for complex Multi-Agent Reinforcement Learning (MARL) systems often remains intractable, we establish a robust rationale for the stability and convergence of our proposed training framework. Its design systematically integrates a series of principles, each targeting a known failure mode in MARL, with convergence anchored in four pillars:

\begin{itemize}
    \item[\textbf{1.}] \textbf{Policy Initialization via Supervised Fine-Tuning (SFT).} A primary challenge in RL lies in the vast and unstructured exploration space, which causes inefficient or divergent training. Our framework addresses this with a curriculum-driven SFT phase. Specifically, it provides a crucial "warm-start" by pre-training each agent on a corpus of high-quality expert policies. Consequently, the MARL optimization process is initialized in a highly structured and effective region of the joint policy space. This approach circumvents the instabilities of \textit{tabula rasa} learning and substantially improves the tractability of subsequent exploration, or as stated in the work, it is essential for ``laying the foundation for collaborative learning in MARL''.

    \item[\textbf{2.}] \textbf{Stable Policy Updates via Group Relative Policy Optimization (GRPO).} The non-stationarity inherent in multi-agent learning where each agent's optimal policy shifts as others learn can destabilize policy updates. The GRPO algorithm directly mitigates this by incorporating a KL-divergence penalty, a core principle of robust policy gradient methods like TRPO and PPO. As shown in Eq.~\ref{eq:grpo_2}, its objective function enforces a trust region for policy updates:
    \begin{equation}
        \max_{\theta} \mathbb{E}_{\mathbf{o}\sim\pi_{\text{old}}}\left[ \frac{\pi_\theta(\mathbf{o}_k)}{\pi_{\text{old}}(\mathbf{o}_k)} A_R^{(k)} - \beta D_{\text{KL}}(\pi_\theta \| \pi_{\text{ref}}) \right]
        \label{eq:grpo_2}
    \end{equation}
    This constraint regularizes learning dynamics by limiting excessive deviations from a trusted reference policy ($\pi_{\text{ref}}$), guaranteeing a monotonic improvement trajectory and fostering training stability.

    \item[\textbf{3.}] \textbf{Dense and Structured Reward Shaping.} MARL systems often face sparse rewards and credit assignment issues, which cause ill-defined optimization landscapes with suboptimal local equilibria. Our framework mitigates this via a dense and multi-faceted reward signal composed of three components: task success ($R_{\text{res}}$), procedural validity ($R_{\text{format}}$), and collaboration quality ($R_{\text{col}}$). This hybrid structure provides a continuous and informative gradient signal, guiding agents toward correct outcomes, valid behaviors and effective cooperation, smoothing the optimization landscape to facilitate gradient-based convergence.

    \item[\textbf{4.}] \textbf{Robustness via Agent Dropout Regularization.} A common failure mode in MARL is inter-agent co-adaptation, where agents develop brittle strategies that are over-specialized to their teammates' specific policies. To address this, we employ agent dropout as a form of structural regularization. By dynamically and stochastically adjusting the communication topology during training, this technique discourages dependencies on any single agent and compels the development of more generalized and robust policies. This enhances the stability of the learned multi-agent equilibrium and ensures convergence to solutions resilient to minor policy perturbations, a fact supported by ablation studies identifying it as the ``most critical regularizer''.
\end{itemize}

In summary, VideoChat-M1’s training convergence is not heuristic but derives from a principled framework design. By systematically addressing initialization (via SFT), update stability (via GRPO), reward-landscape tractability (via dense rewards) and robust generalization (via agent dropout), the framework holistically mitigates common MARL instabilities, guiding the agent system toward a stable and high-performance collaborative pipeline.

\section*{A.6 More Implementation Details}

\paragraph{Training Setup.}
We employ the AdamW~\cite{adamw} optimizer with a learning rate of $1\text{e-}7$ and a global batch size of 8. The training process is distributed across 8 NVIDIA A100 80G GPUs, utilizing the DeepSpeed stage 2 combined with \textbf{Flash Attention~\cite{dao2023flashattention2} and bfloat16 precision} to accelerate multi-GPU training and optimize memory efficiency. The gradient accumulation step is set to 2. 
Our agent team consists of four backbone models: Qwen3-8B, Qwen3-4B, Qwen2.5-7B, and Qwen2.5-3B. During training, the temperature is set to 1 for each agent to facilitate exploration, while the KL penalty coefficient $\beta$ is set to $1\text{e-}5$. We set the maximum prompt length to 1024 tokens and the maximum generation length to 1024 tokens. The multi-agent interaction is limited to a maximum of 5 turns. Specific prompts are provided in Appendix A.2. Additionally, we apply agent dropout to enhance the model's robustness.

\paragraph{LoRA Setting.}
As reported in Tab 8 of the submitted manuscript, we implement Low-Rank Adaptation (LoRA) using the Hugging Face \texttt{peft} library. The LoRA adapters are configured with a rank $r=8$, a scaling factor $\alpha=16$, and a dropout rate of $0.05$. We adjust the learning rate specifically for LoRA training to $2\text{e-}6$. Except for these specific adjustments, all other hyperparameters remain consistent with the full fine-tuning configuration described above.

\paragraph{Optimization Strategy.}
We adopt Group Relative Policy Optimization (GRPO) as our reinforcement learning algorithm. GRPO is selected for its suitability in scenarios involving optimization from a group of candidate outputs. By normalizing rewards against the team's average performance, GRPO provides a stable learning signal for each individual agent, aligning naturally with our multi-agent collaborative generation paradigm.

\paragraph{SFT Data Construction.}
To enable efficient Supervised Fine-Tuning (SFT), we construct a filtered dataset derived from successful interaction trajectories. As illustrated in our pipeline (see Figure~\ref{fig:sft}), tools, questions, and options are input into the Agent Team. Following the Collaborative Policy Planning Process (CPP), a final answer is generated. We retain a trajectory only if: (1) at least one agent provides the correct answer, and (2) the initial plan remains unchanged throughout the process. We collect 2,000 such initial plans per task. This filtering strategy reduces unnecessary self-correction steps and significantly improves computational efficiency.

\paragraph{Evaluation Setup.}
For evaluation, all LLMs use a temperature of 0 to ensure deterministic outputs. The agent group composition remains consistent with the training phase (Qwen3-8B, Qwen3-4B, Qwen2.5-7B, and Qwen2.5-3B), totaling approximately 22B parameters. 
Our reported inference latency (19.8s) is achieved with 4 A100 80G GPUs via parallel processing and bfloat16 precision: (1) we implement parallel processing across the Policy Generation, Execution, and Communication stages, enabling concurrent reasoning and tool invocation across agents (instead of sequential processing); (2) we enforce strict token constraints during reasoning to prompt concise rationales, significantly reducing the decoding overhead. 
This evaluation can be run on only one A100 80G GPU for each task with partial parallel processing, with about 38.9s per video with 67G VRAM. However, a single A100 80G GPU is insufficient for inference on an MLLM of 72B+ parameters to handle long videos with 100+ sampled frames. When all invoked tools are exhausted or the maximum number of iterations is reached without QA consensus, we directly use Qwen3-8B to generate a summarized answer using the memory of all agents.

\paragraph{Tool Configurations.}
We tailor the tool set and underlying models for specific benchmarks to maximize performance. The standard tool library includes: \textit{Global Sampling, Video Retrieval, Time Stamp Retrieval, Rough Browser, Fine Browser,} and \textit{Grounding Tool}.
For image retrieval, we use ViT-CLIP-B/16 (86M). For video retrieval, we use ASP-CLIP (95M).

\begin{itemize}
    \item \textbf{General Video QA (LongVideoBench, Video-MME, MLVU, Video Holmes, MMR-V):} 
    We utilize the standard tool library. The \textit{Browser} model is instantiated with Qwen2.5-VL-7B, and \textit{Grounding Tool} employs Eagle2.5-8B. The grounding tool is invoked with relatively low frequency.
    The total parameter count for the toolset is approximately 37B.
    
    \item \textbf{Video MMMU:} 
    The configuration largely follows that of the General Video QA setup, except that the \textit{Browser} model is upgraded to Qwen3VL-8B-Instruct to handle higher domain-specific demands. The total parameter count is approximately 37B.
    
    \item \textbf{Video VSIBench (Spatial Tasks):} 
    We introduce a specialized \textit{Spatial Tool} powered by InternVL3.5-8B. For spatial queries, the model autonomously selects between the \textit{Browser} (Qwen2.5-VL-7B) or the \textit{Spatial Tool} for answer generation. The total parameter count is approximately 37B.
     
    \item \textbf{Charades-STA (Temporal Grounding):} 
    The model dynamically chooses between the \textit{Browser} and the \textit{Grounding Tool}. Input videos are processed at 2 FPS. If the \textit{Video Retrieval} tool is invoked, the retrieved video clip is subsequently fed into the model for fine-grained grounding. The total number of parameters is approximately 37B. 
    For this dataset, we select up to three consecutive video clips. We first select the clip with the highest similarity; if the similarity of an adjacent clip with the key info exceeds 0.35, we include it as well. This prevents the situation where the answer's grounding time exceeds the duration of the retrieved clip. Additionally, it narrows the retrieval interval and eliminates redundant information, thereby improving performance

\end{itemize}

\begin{table}

    \centering
    \small
    \setlength{\tabcolsep}{1mm}
    \renewcommand{\arraystretch}{0.85}
    \begin{tabular}{c|c|c}
        \toprule
        \textbf{Spatial Tool} &  \textbf{Baseline} & \textbf{VideoChat-M1}\\  \midrule
        InternVL3.5-8B & 56.3 & 71.9 \\
        Qwen2.5VL-7B &  35.9 &  70.1\\
        \bottomrule
    \end{tabular}
    \vspace{-0.75em}
    
    \caption{Tool Reliance Ablation on VSIBench}
    \label{tab:tool_reli}

\vspace{-1em}

\end{table}

\paragraph{Tool Reliance Ablation:}
To demonstrate that the effectiveness of VideoChat-M1 originates from our Collaborative Policy Planning (CPP) framework rather than reliance on specific SOTA tools, we conducted an additional ablation study on VSIBench (see Tab~\ref{tab:tool_reli}).
Specifically, we replaced the specialized 'Spatial Tool' (InternVL) with the general-purpose Qwen2.5-VL-7B. Remarkably, even with this generic backbone, our method retains SOTA performance. It continues to outperform the massive InternVL-3.5-241B, achieving a 34.2\% improvement over the baseline. This confirms that our MARL-driven planning paradigm delivers substantial gains, independent of the specific tools employed.

\end{document}